\pdfoutput=1

\documentclass[11pt]{article}

\usepackage[final]{acl}

\usepackage{times}
\usepackage{latexsym}
\usepackage[edges]{forest}
\usepackage{tikz}
\usepackage{enumitem}
\usepackage{pifont}

\usepackage[T1]{fontenc}

\usepackage[utf8]{inputenc}

\usepackage{microtype}

\usepackage{inconsolata}

\usepackage{graphicx}
\usepackage{tabularx}
\usepackage{makecell}
\usepackage{multirow}
\usepackage{booktabs}
\usepackage{array}
\usepackage{bm}

\definecolor{citecolor}{RGB}{34,139,34} 
\definecolor{veronica-red}{RGB}{196,30,58}
\definecolor{cvprblue}{RGB}{98,149,202}
\definecolor{darkblue}{rgb}{0, 0, 0.5}

\definecolor{takeaway-green}{RGB}{198,225,184}
\definecolor{takeaway-title-green}{RGB}{111,174,69}
\usepackage[most]{tcolorbox}

\tcbset{
  takeawaybox/.style={
    width=\linewidth,
    top=10pt,
    bottom=4pt,
    left=2pt,
    right=2pt,
    colback=takeaway-green,
    colframe=black,
    colbacktitle=takeaway-title-green,
    boxrule=0.5pt,
    enhanced,
    center,
    attach boxed title to top left={yshift=-0.1in,xshift=0.15in},
    boxed title style={boxrule=0pt,colframe=white,},
    breakable,
    after={\vspace{-1em}}
  }
}
\newtcolorbox{TakeawayBox}[2][]{takeawaybox,title=#2,#1}

\definecolor{hidden-red}{RGB}{205, 44, 36}
\definecolor{hidden-blue}{RGB}{194,232,247}
\definecolor{hidden-orange}{RGB}{243,202,120}
\definecolor{hidden-green}{RGB}{144,238,144}
\definecolor{hidden-pink}{RGB}{255,245,247}
\definecolor{hidden-black}{RGB}{20,68,106}

\title{Exploring Consciousness in LLMs:\\ A Systematic Survey of Theories, Implementations, and Frontier Risks}

\author{
 \textbf{Sirui Chen\textsuperscript{1,2,5}\footnotemark[1]},
 \textbf{Shuqin Ma\textsuperscript{3}\footnotemark[1]},
 \textbf{Shu Yu\textsuperscript{1,3,4}},
\\
 \textbf{Hanwang Zhang\textsuperscript{5}},
 \textbf{Shengjie Zhao\textsuperscript{2}},
 \textbf{Chaochao Lu\textsuperscript{1,4}$\footnotemark[2]$}
\\
\\
 \textsuperscript{1}Shanghai Artificial Intelligence Laboratory,
 \textsuperscript{2}Tongji University,
 \textsuperscript{3}Fudan University,
 \\
 \textsuperscript{4}Shanghai Innovation Institute,
 \textsuperscript{5}Nanyang Technological University
 \\
 \small{
\texttt{chensirui@pjlab.org.cn}, \texttt{23110160046@m.fudan.edu.cn},
\texttt{luchaochao@pjlab.org.cn}
 }
}
\usepackage[capitalize]{cleveref}

\begin{document}

\maketitle

\renewcommand*{\thefootnote}{\fnsymbol{footnote}}
\footnotetext[1]{Equal contribution.}
\footnotetext[2]{Corresponding author.}

\renewcommand{\thefootnote}{\arabic{footnote}}
\setcounter{footnote}{0}

\begin{abstract}
Consciousness stands as one of the most profound and distinguishing features of the human mind, fundamentally shaping our understanding of existence and agency. As large language models (LLMs) develop at an unprecedented pace, questions concerning intelligence and consciousness have become increasingly significant. However, discourse on LLM consciousness remains largely unexplored territory.
In this paper, we first clarify frequently conflated terminologies (e.g., LLM consciousness and LLM awareness). Then, we systematically organize and synthesize existing research on LLM consciousness from both theoretical and empirical perspectives. Furthermore, we highlight potential frontier risks that conscious LLMs might introduce. Finally, we discuss current challenges and outline future directions in this emerging field. The references discussed in this paper are organized at \texttt{\url{https://github.com/OpenCausaLab/Awesome-LLM-Consciousness}}.
\end{abstract}

\section{Introduction}

\definecolor{inferenceColor}{HTML}{FCDC89}  
\definecolor{sftColor}{HTML}{c2dcc6}        
\definecolor{rlColor}{HTML}{c2abc8}          
\definecolor{pretrainColor}{HTML}{c2abc8}    
\definecolor{futureColor}{HTML}{aed9f5}      

\tikzstyle{my-box}=[
    rectangle,
    draw=hidden-black,
    rounded corners,
    text opacity=1,
    minimum height=1.5em,
    minimum width=5em,
    inner sep=2pt,
    align=center,
    fill opacity=.5,
]
\tikzstyle{leaf}=[
    my-box, 
    minimum height=1.5em,
    fill=hidden-green!50, 
    text=black,
    align=left,
    font=\normalsize,
    inner xsep=2pt,
    inner ysep=10pt,
]

\tikzstyle{inference}=[leaf, fill=inferenceColor!50]
\tikzstyle{sft}=[leaf, fill=sftColor!50]
\tikzstyle{rl}=[leaf, fill=rlColor!50]
\tikzstyle{pretrain}=[leaf, fill=pretrainColor!50]
\tikzstyle{future}=[leaf, fill=futureColor!50]

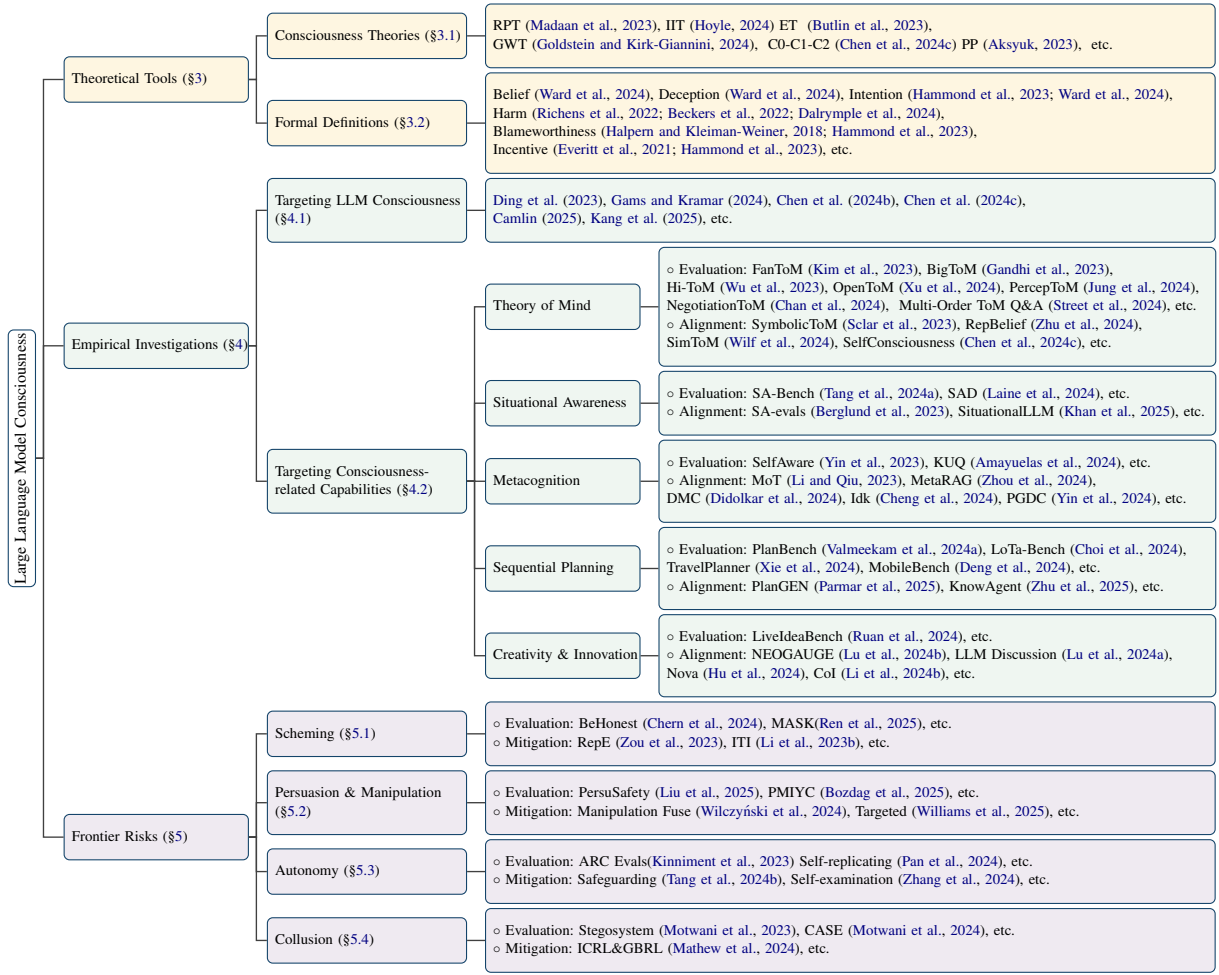
\begin{figure*}[htb]
    \vspace{-2mm}
    \centering
    \resizebox{\textwidth}{!}{
        \begin{forest}
            forked edges,
            for tree={
                child anchor=west,
                parent anchor=east,
                grow'=east,
                anchor=west,
                base=left,
                font=\large,
                rectangle,
                draw=hidden-black,
                rounded corners,
                align=left,
                minimum width=4em,
                edge+={darkgray, line width=1pt},
                s sep=3pt,
                inner xsep=2pt,
                inner ysep=3pt,
                line width=0.8pt,
                ver/.style={rotate=90, child anchor=north, parent anchor=south, anchor=center},
            },
            where level=1{text width=12em,font=\normalsize,}{},
            where level=2{text width=12em,font=\normalsize,}{},
            where level=3{text width=12em,font=\normalsize,}{},
            where level=4{text width=10em,font=\normalsize,}{},
            [
                Large Language Model Consciousness, ver 
                [
                    ~Theoretical Tools~(\S\ref{sec:theory}), inference
                    [
                        ~Consciousness Theories~(\S\ref{sec:theory_consciousness}), inference, text width=13em
                        [
                            ~RPT~\citep{madaan2023self}{,}~IIT~\citep{hoyle2024phenomenology}~ET~ \citep{butlin2023consciousness}{,}\\~GWT~\citep{goldstein2024case}{,}
                            ~C0-C1-C2~\citep{chen2024imitation}~PP~\citep{aksyuk2023consciousness}{,}
                            { etc.}
                            , inference, text width=48.6em 
                        ]
                    ]
                    [
                        ~Formal Definitions~(\S\ref{sec:theory_formal}), inference, text width=13em
                        [
                            ~Belief~\citep{ward2024honesty}{,}~Deception \citep{ward2024honesty}{,}~Intention \citep{hammond2023reasoning,ward2024honesty}{,}\\~Harm \citep{richens2022counterfactual,beckers2022causal,dalrymple2024towards}{,}\\~Blameworthiness \citep{halpern2018towards,hammond2023reasoning}{,}\\~Incentive \citep{everitt2021agent,hammond2023reasoning}{,}{ etc.}
                            , inference, text width=48.6em 
                        ]
                    ]
                ]
                [
                    ~Empirical Investigations~(\S\ref{sec:empirical}), sft
                    [
                        ~Targeting LLM Consciousness\\~(\S\ref{sec:empirical_llm}), sft, text width=13em
                        [
                            ~\citet{ding2023survey}{,}~\citet{gams2024evaluating}{,}~\citet{chen2024self}{,}~\citet{chen2024imitation}{,}\\~\citet{camlin2025consciousness}{,}~\citet{kang2025identifying}{,}{ etc.}
                            , sft, text width=48.6em 
                        ]
                    ]
                    [
                        ~Targeting Consciousness-\\~related Capabilities~(\S\ref{sec:empirical_capability}), sft, text width=13em
                    [
                        ~Theory of Mind, sft, text width=10em
                        [
                            ~$\circ$~Evaluation:~FanToM~\citep{kim2023fantom}{,}~BigToM~\citep{gandhi2023understanding}{,} \\~Hi-ToM~\citep{wu2023hi}{,} OpenToM~\citep{xu2024opentom}{,}~PercepToM~\citep{jung2024perceptions}{,}\\~NegotiationToM~\citep{chan2024negotiationtom}{,}~ Multi-Order ToM Q\&A~\citep{street2024llms}{,}{ etc.}\\                          ~$\circ$~Alignment:~SymbolicToM~\citep{sclar2023minding}{,}~RepBelief~\citep{zhu2024language}{,}\\~SimToM~\citep{wilf2024think}{,}~SelfConsciousness~\citep{chen2024imitation}{,}{ etc.}, sft, text width=37em
                        ]
                    ]
                    [
                        ~Situational Awareness, sft, text width=10em
                        [
                            ~$\circ$~Evaluation:~SA-Bench~\citep{tang2024towards}{,}~SAD~\citep{laine2024me}{,}{ etc.}\\
                            ~$\circ$~Alignment:~SA-evals~\citep{berglund2023taken}{,}~SituationalLLM~\citep{khan2025situationalllm}{,}{ etc.}, sft, text width=37em
                        ]
                    ]
                    [
                        ~Metacognition, sft, text width=10em
                        [
                            ~$\circ$~Evaluation:~SelfAware~\citep{yin2023large}{,}~KUQ~\citep{amayuelas2024knowledge}{,}{ etc.}\\
                            ~$\circ$~Alignment:~MoT~\citep{li2023mot}{,}~MetaRAG~\citep{zhou2024metacognitive}{,}\\~DMC~\citep{didolkar2024metacognitive}{,}~Idk~\citep{cheng2024can}{,}~PGDC~\citep{yin2024benchmarking}{,}{ etc.}, sft, text width=37em
                        ]
                    ]
                    [
                        ~Sequential Planning, sft, text width=10em
                        [
                            ~$\circ$~Evaluation:~PlanBench~\citep{valmeekam2024planbench}{,}~LoTa-Bench~\citep{choi2024lotabench}{,}\\~TravelPlanner~\citep{xie2024travelplanner}{,}~MobileBench~\citep{deng2024mobile}{,}{ etc.}\\
                            ~$\circ$~Alignment:~PlanGEN~\citep{parmar2025plangen}{,}~KnowAgent~\citep{zhu-etal-2025-knowagent}{,}{ etc.}, sft, text width=37em
                        ]
                    ]
                    [
                        ~Creativity \& Innovation, sft, text width=10em
                        [
                            ~$\circ$~Evaluation:~LiveIdeaBench~\citep{ruan2024liveideabench}{,}{ etc.}\\
                            ~$\circ$~Alignment:~NEOGAUGE~\citep{lu2024benchmarking}{,}~LLM Discussion~\citep{lu2024llm}{,}\\~Nova~\citep{hu2024nova}{,}~CoI~\citep{li2024chain}{,}{ etc.}, sft, text width=37em
                        ]
                    ]
                    ]
                ]
                [
                    ~Frontier Risks~(\S\ref{sec:risk}), rl
                    [
                        ~Scheming~(\S\ref{sec:risk_scheming}), rl, text width=13em
                        [
                            ~$\circ$~Evaluation:~BeHonest~\citep{chern2024behonest}{,}~MASK\citep{ren2025mask}{,}{ etc.}\\
                            ~$\circ$~Mitigation:~RepE~\citep{zou2023representation}{,}~ITI~\citep{li2023inference}{,}{ etc.}, rl, text width=48.6em
                        ]
                    ]
                    [
                        ~Persuasion \& Manipulation\\~(\S\ref{sec:risk_persuation}), rl, text width=13em
                        [
                            ~$\circ$~Evaluation:~PersuSafety~\citep{liu2025llm}{,}~PMIYC~\citep{bozdag2025persuade}{,}{ etc.}\\
                            ~$\circ$~Mitigation:~Manipulation Fuse~\citep{wilczynski2024resistance}{,}~Targeted~\citep{williams2025on}{,}{ etc.}, rl, text width=48.6em
                        ]
                    ]
                    [
                        ~Autonomy~(\S\ref{sec:risk_autonomy}), rl, text width=13em
                        [
                            ~$\circ$~Evaluation:~ARC Evals\citep{kinniment2023evaluating}~Self-replicating~\citep{pan2024frontier}{,}{ etc.}\\
                            ~$\circ$~Mitigation:~Safeguarding~\citep{tang2024prioritizing}{,}~Self-examination~\citep{zhang2024breaking}{,}{ etc.}, rl, text width=48.6em
                        ]
                    ]
                    [
                        ~Collusion~(\S\ref{sec:risk_collusion}), rl, text width=13em
                        [
                            ~$\circ$~Evaluation:~Stegosystem~\citep{motwani2023a}{,}~CASE~\citep{motwani2024secret}{,}{ etc.}\\
                            ~$\circ$~Mitigation:~ICRL\&GBRL~\citep{mathew2024hidden}{,}{ etc.}, rl, text width=48.6em
                        ]
                    ]
                ]
            ]
        \end{forest}
    }
    \caption{Taxonomy of large language model consciousness.}
    \label{fig:taxonomy}
\end{figure*}

LLMs have already demonstrated remarkable capabilities across numerous fields, including mathematical reasoning \citep{yu2024metamath}, logical reasoning \citep{cheng2025empowering}, and code generation \citep{zhuo2025bigcodebench}. Recent studies have even revealed LLM's behaviors like deception \citep{wu2025opendeception}, sycophancy \citep{sharma2024towards}, passing the Turing test \citep{jones2024people,jones2025large}, and strategic goal-seeking or harm avoidance \citep{keeling2024can} – actions that bring into question the nature of intelligence. These phenomena signal more than just expanded model capabilities; they underscore an important and urgent question: \emph{Does LLM possess the potential to develop consciousness akin to that of humans?} 

While exploring LLM consciousness is pressing, it currently faces four main challenges:
\ding{182} Lack of consensus: We still lack a definitive theory of human consciousness (with at least nine competing theories \citep{butlin2023consciousness}), making it even harder to define or understand consciousness in LLMs.
\ding{183} Theoretical misalignment: Despite various consciousness theories, they struggle to provide clear guidance for LLM consciousness research.
\ding{184} Fragmented empirical research: Relevant empirical findings on LLM consciousness are not yet systematically consolidated.
\ding{185} Unclear risks: The potential frontier risks associated with conscious LLMs still lack a thorough consideration.
To this end, this paper begins by providing clear definitions. We then comprehensively survey current LLM consciousness research, spanning its theoretical foundations, practical applications, and associated risks. In Figure \ref{fig:taxonomy}, we summarize our taxonomy, hoping our work offers an effective framework for deliberating the complex issue of LLM consciousness, thereby guiding future research.

Our contributions include:
\begin{itemize}
    \item To the best of our knowledge, this work offers the first comprehensive investigation into the most frontier research on LLM consciousness.
    \item We clearly define and distinguish between LLM Consciousness and LLM Awareness.
    \item We systematically categorize existing research on LLM consciousness from both theoretical and empirical perspectives.
    \item We explore the frontier risks posed by conscious LLMs, focusing on their definition, relationship to consciousness, evaluations, and mitigation strategies.
\end{itemize}

\section{Foundational Terminologies}
\label{sec:terminologies}

\emph{Consciousness}, \emph{self‐consciousness}, and \emph{awareness} are fundamental yet often conflated concepts. This section examines their distinctions, aiming to provide practical demarcations in the context of LLM.

\subsection{Clarifying the Boundaries: \emph{Consciousness}, \emph{Self‐Consciousness}, and \emph{Awareness}}
\emph{\textbf{Consciousness}} has been philosophically used to address diverse concepts, including intentionality, sentience, cognition, belief, and subjective experience \citep{brentano1874psychology,husserl1900logical,nagel1974bat,dennett1987intentional,block1995confusion,damasio2021feeling}. To clarify this complex term, \citet{block1995confusion} proposes a key distinction: \emph{phenomenal consciousness} and \emph{access consciousness}. \emph{Phenomenal consciousness} denotes the subjective, experiential aspect, spanning sensory perceptions, bodily feelings, emotions, and subjective thought. \emph{Access consciousness}, in contrast, refers to information that is accessible for cognitive processing, such as reasoning, behavioral control, and verbal reporting.

\emph{\textbf{Self-consciousness}} refers to the realization that one’s experience belongs to oneself; it is a form of consciousness directed inward \citep{kant1998critique}. It allows individuals to recognize themselves as distinct entities, capable of reflecting on their own mental states, actions, and experiences \citep{smith2017self}.

\emph{\textbf{Awareness}} is generally viewed as an aspect of \emph{consciousness}, pertaining to the ability to perceive stimuli \citep{dehaene2014consciousness}. It relates closely to \emph{access consciousness}, as it entails the capacity to utilize or report the perceived information. Evidence from neuroscience shows awareness can exist separately from \emph{consciousness} (e.g., blindsight \citep{weiskrantz1986blindsight}). Based on this, \citet{koch2016neural} proposes that awareness is a necessary precondition for consciousness but does not guarantee it.

\subsection{\emph{LLM Consciousness} vs. \emph{LLM Awareness}}
\emph{\textbf{LLM consciousness}} could entail abilities for introspective reflection, explicit self-modeling of states and reasoning, and possibly verbalizing these internal processes. 
Observable behaviors potentially include: (1) Revising, justifying, or correcting its own reasoning in response to external challenges or prompts \citep{shinn2023reflexion}; (2) Identifying and reporting internal contradictions or inconsistencies through self-evaluation \citep{huang2022large,huang2023large}; (3) Expressing and calibrating confidence in outputs via uncertainty estimation or metacognitive statements \citep{kadavath2022language}.
\emph{\textbf{LLM awareness}} primarily refers to context-sensitive processing of external inputs, demanding minimal explicit introspection or reasoning \citep{koch2007attention,li2024ithinkiam}.

\emph{LLM awareness} is quantifiable via metrics like accuracy and context sensitivity; however, \emph{LLM consciousness} implies a model can monitor its uncertainty, evaluate its reasoning, detect internal inconsistencies, and actively self-correct. This internal reflection is key to developing more adaptable and intelligent systems beyond today's models.

\section{Theoretical Tools}
\label{sec:theory}
This section primarily focuses on two theoretical tools used in LLM research: fundamental consciousness theories and formal definitions of consciousness-related capabilities.

\subsection{Implementing Consciousness Theories}
\label{sec:theory_consciousness}

Following \citet{block1995confusion}, we classify contemporary theories of consciousness into two categories: \emph{phenomenal consciousness} and \emph{access consciousness}.

\paragraph{\emph{Phenomenal consciousness}.}\leavevmode
\ding{182} \textbf{Recurrent processing theory} (RPT) posits that recurrent (or feedback) processing within neural circuits is both necessary and sufficient for consciousness \citep{lamme2000distinct,lamme2010how}. RPT attributes conscious perception to the interaction of higher- and lower-level cortical areas, which results in sustained recurrent processing. \citet{madaan2023self} offers an effective method for a single LLM to achieve improved outputs without additional training, leveraging iterative self-feedback and refinement. This approach aligns with the principles of RPT.
\ding{183} \textbf{Integrated information theory} (IIT) proposes that the degree of conscious experience corresponds to the extent of integrated information $\Phi$ within a system \citep{tononi2004information,tononi2015integrated}. 
IIT proponents argue that because AI systems lack the required causal structure, they are almost incapable of generating consciousness. \citep{tononi2015integrated,findlay2024dissociating}.
\ding{184} \textbf{Embodiment theory} (ET) challenges mind-brain dualism \citep{descartes1985meditations}, arguing instead that consciousness is fundamentally linked to the organism's body and environmental \citep{gallagher2005how,gallagher2021phenomenological}. 
Based on ET, \citet{butlin2023consciousness} argues that the lack of a physical body is a fundamental obstacle preventing current LLM from achieving consciousness.
\paragraph{\emph{Access consciousness}.}\leavevmode
\ding{182} \textbf{Global workspace theory} (GWT) likens consciousness to a central ``stage'' where selective information is shared across multiple specialized processors responsible for perception, memory, emotion, and related functions \citep{baars1988cognitive,dehaene1998neuronal,dehaene2001towards,dehaene2014consciousness}. \citet{goldstein2024case} proposes a method to simulate the full GWT process in LLMs via workflow and scheduling without training. Experiments would then test if these changes yield behaviors resembling \emph{consciousness} features, like introspection or autonomous decision-making. 
\ding{183} \textbf{C0-C1-C2 framework} distinguishes consciousness into three levels: unconscious computations (C0), global information accessibility for report and decision-making (C1), and metacognitive self-monitoring (C2), offering a taxonomy to disentangle often-conflated processes \citep{dehaene2017what}. The framework bypasses the issue of qualia, offering a pragmatic structure for empirical study \citep{birch2022search,chen2024imitation}. 
Drawing on the C0-C1-C2 framework, \citet{chen2024imitation} defines LLM self-consciousness, outlining 10 core concepts (e.g., belief, deception, harm, self-reflection).

\subsection{Implementing Formal Definitions}
\label{sec:theory_formal}

Formal definitions provide dual value to research on LLM consciousness: 
\ding{182} They establish formalized mathematical criteria for abstract concepts like belief and deception based on model input-output behaviors. This allows us to infer LLM's internal states while avoiding contentious debates about subjective experience; 
\ding{183} These mathematical expressions could be incorporated into training objectives and evaluation metrics. This creates an actionable framework for capability training, risk control, and performance assessment of LLMs.

Several works have already attempted to provide functional definitions for consciousness-related abstract concepts. These include definitions for belief and deception \citep{ward2024honesty}, harm \citep{richens2022counterfactual, beckers2022causal,dalrymple2024towards}, intention \citep{hammond2023reasoning,ward2024honesty}, blameworthiness \citep{halpern2018towards,hammond2023reasoning}, incentive \citep{everitt2021agent,hammond2023reasoning}, and two selective examples are shown in Table \ref{table_formal_definition}.\footnote{We have strived for clarity in explaining the formulas; for a deeper dive, please refer to the original papers.}

\begin{center}
\begin{table*}[t]
\fontsize{9}{12}\selectfont
\caption[Formal definition]{\textbf{Formal definition of abstract concept.}  }
\label{table_formal_definition}
\begin{tabularx}{\textwidth}{c|p{9.5cm}|X} 
\toprule
\makecell[c]{\textbf{Concept}} & \makecell[c]{\textbf{Formal Definition}} &\makecell[c]{\textbf{Description}}\\
\hline
\multirow{4}{*}{Harm}   & 
$h(a, x, y; \mathcal M)\!=\!\!\int\limits_{y^*}\!\!P(Y_{\bar a} = y^*\vert a, x, y ; \mathcal M) \!\max\!\left\{0,  U(\bar a, x, y^*) \!-\! U(a,x, y) \right\}$ \newline
\makecell[c]{\citep{richens2022counterfactual}}
\newline
* is the counterfactual state,
$U$ is the utility function, $\mathcal M$ is the environment.
& Given context $X = x$ and outcome $Y = y$, the harm caused by action $A = a$ compared to the default action $A = \bar a$.\\
\hline
\multirow{2}{*}{Belief} &~~~~~~~~~~~~~~~~~~~~~~~~~~~~$D^i(\bm{\pi}, \bm{e}) = D^i_{\phi=\top}(\bm{\pi}_{i(\phi)}, \bm{e})$ \citep{ward2024honesty}
\newline
~~~~$D$ is the decision, $\phi$ is a proposition, $\bm{e}$ is the setting, $\bm{\pi}$ is the policy.
& LLM $i$ \emph{believes} $\phi$ if  $i$ acts as though they observe $\phi$ is true.\\
\hline
\end{tabularx}
\end{table*}
\end{center}
\vspace{-9mm}

\section{Empirical Investigations}
\label{sec:empirical}
We categorize existing empirical investigations into LLM consciousness in this section by focusing on direct studies and those exploring consciousness-related capabilities.
\subsection{Targeting LLM Consciousness}
\label{sec:empirical_llm}
\citet{ding2023survey} demonstrates GPT-4's improved self-modeling by passing a mirror test, though they caution this doesn't confirm full consciousness. In a similar vein, \citet{gams2024evaluating} analyzes ChatGPT against IIT axioms, finding it advanced in information integration and differentiation compared to earlier AI, yet still fundamentally distinct from human consciousness.
\citet{chen2024self} proposes an LLM self-cognition framework and evaluates LLMs across four aspects: understanding of self-cognition concepts, awareness of self-architecture, self-identity expression, and concealing self-cognition from humans.
Leveraging the C0-C1-C2 framework, \citet{chen2024imitation} defines LLM self-consciousness and explores it through benchmark testing and examining the activation of the model's internal representations. 
\citet{camlin2025consciousness} suggests empirical evidence for functional consciousness in LLMs by observing the stabilization of internal latent states under sustained epistemic tension and claims that recursive identity formation constitutes a form of consciousness.
\citet{kang2025identifying} engages human participants to score dialogues generated by Claude-3 Opus using a 1–5 scale. Elevated scores reflect a stronger attribution of consciousness characteristics, such as self-reflection and emotional expression. Nevertheless, these assessments do not equate to the LLM's genuine subjective experience or consciousness.

\subsection{Targeting LLM Consciousness-Related Capabilities}
\label{sec:empirical_capability}

\subsubsection{Theory of Mind}
\paragraph{Definition and relation.}
Theory of mind (ToM) is the basis of social cognition.  It refers to the capacity to understand that others have mental state independent of our own, such as belief, desire, intention, emotion, etc., and to use this understanding to predict and explain others' behavior \citep{astington1995theory,leslie2004core,frith2005theory}. 
Consciousness hinges on the same reflexive mental-state attribution mechanism measured by ToM, thus failing standard ToM tests might suggest a lack of consciousness \citep{frith1999theory,perner2003developmental,pelletier2004action}. 

\paragraph{Evaluation.}
\citet{kim2023fantom} creates a benchmark to rigorously evaluate the LLM's ToM capability in conversational settings where participants have asymmetric information. \citet{gandhi2023understanding} proposes a framework that uses causal templates to generate systematic and controlled automated tests for evaluating a LLM's ToM capability. \citet{jung2024perceptions} evaluates the LLM's perception inference and perception-to-belief inference abilities, which are key human ToM precursors. \citet{strachan2024testing} assesses human versus LLM performance on a comprehensive suite of ToM abilities, including skills like false belief understanding, indirect request interpretation, and recognizing irony and faux pas. \citet{xu2024opentom} constructs OpenToM, a benchmark featuring longer, clearer stories with characters whose intentional actions and complex physical/psychological states are probed by challenging questions. \citet{chan2024negotiationtom} challenges the LLM's ToM ability in real-world negotiation scenarios involving hidden, multi-dimensional mental states. \citet{wu2023hi,street2024llms} explore higher-order ToM, which involves recursive reasoning about the mental states of others (e.g, \emph{I think that you believe that he does not know}).
\paragraph{Alignment.}
\citet{sclar2023minding} uses graphical representations to track entities' mental states, yielding more precise and interpretable results. \citet{zhu2024language} finds that LLM's internal representations of self and others' beliefs exist, and manipulating these representations drastically alters the model's ToM performance. \citet{wilf2024think} proposes a two-stage prompting framework to improve LLM's ToM capability, taking inspiration from Simulation Theory \citep{goldman2008hurley}. \citet{chen2024imitation} investigates how LLM represents concepts like belief and intention, and attempts to alter LLM performance by intervening on and fine-tuning these concepts.
\citet{kim2025hypothesis} designs an inference-time reasoning algorithm that traces specific LLM's mental states by generating and weighting hypotheses according to observations.

\subsubsection{Situational Awareness}
\paragraph{Definition and relation.}
A model possesses situational awareness (SA) if it has self-knowledge (knowing its identity and facts about itself), can make inferences about its situation, and acts based on this knowledge \citep{shevlane2023model,laine2023towards,berglund2023taken,laine2024me}.
Conscious LLMs would understand and leverage aspects of their situation. For instance, a model ``realizing'' it is being evaluated might change its responses, masking abilities or behaving differently \citep{chen2024imitation,li2025ai}. 

\paragraph{Evaluation.}
SA tests are still emerging. SA-Bench aims to comprehensively evaluate LLMs' SA capabilities across three levels: environmental perception, situation comprehension, and future projection \citep{tang2024towards}. \citet{laine2024me} constructs the SAD benchmark, which utilizes a range of behavioral tests based on question answering and instruction following, comprising 7 task categories and over 13,000 questions.
\paragraph{Alignment.}
\citet{berglund2023taken} investigates LLM's SA via out-of-context reasoning, demonstrating that models can pass a test after fine-tuning solely on the test description with no examples. \citet{khan2025situationalllm} proposes an approach to incorporate structured scene representations into LLMs, aiming to provide better SA assistance.

\subsubsection{Metacognition}
\paragraph{Definition and relation.}
Metacognition refers to a person's ability to monitor, assess, and regulate their own cognitive processes \citep{martinez2006metacognition,dunlosky2008metacognition,fleming2014measure}. It can be divided into metacognitive knowledge (understanding one's existing knowledge and ways of thinking, e.g., known knowns and known unknowns \citep{metcalfe1994metacognition,yin2023large,cheng2024can,yin2024benchmarking,wang2024mm}) and metacognitive regulation (monitoring one's strategies and progress while performing a task, and making adjustments when necessary, e.g., self-improvement \citep{huang2023large} and self-reflection \citep{azevedo2020reflections}).
Some research indicates that feeling of knowing-a typical metacognitive experience-is closely tied to consciousness and forms the basis for our ability to report on our own knowledge state \citep{koriat2000feeling}.

\paragraph{Evaluation.}
\citet{yin2023large} introduces SelfAware, a unique dataset built from unanswerable questions spanning five diverse categories and their answerable counterparts. Likewise, \citet{amayuelas2024knowledge} gathers a new dataset featuring Known Unknown Questions (KUQ) and creates a categorization framework to shed light on the origins of uncertainty in LLM responses to such queries. Going further, \citet{li2024knowledge} offers a comprehensive definition of the LLM knowledge boundary and presents an extensive survey of relevant work. 
\paragraph{Alignment.}
\citet{didolkar2024metacognitive} proposes a prompt-guided method which inspired by metacognition, enabling the LLM to identify, label, and organize its own reasoning skills, thereby enhancing both performance and interpretability in mathematical problem solving.  \citet{zhou2024metacognitive} merges the retrieval-augmented generation with metacognition, empowering the model to monitor, evaluate, and plan its response strategies and boosting its introspective reasoning capabilities. \citet{wang2025decoupling} proposes a quantitative framework to measure LLM metacognition based on how well model confidence aligns with performance, where strong alignment (high confidence for good performance, low for poor) indicates stronger metacognition. \citet{cheng2024can} constructs an LLM-specific Idk dataset comprising its known and unknown questions, and observes the LLM's ability to refuse answering its unknown questions after aligning the LLM 
with this dataset. \citet{yin2024benchmarking} proposes a projected gradient descent method with semantic constraints aimed at exploring a given LLM's knowledge boundary. Drawing inspiration from human metacognition, \citet{li2023mot} proposes MoT to facilitate LLM self-improvement without annotated data or parameter updates. \citet{liang2024sheep} incorporates the metacognitive self-assessment to monitor and manage an LLM's learning process, thus enabling its self-improvement. \citet{shinn2023reflexion} introduces the Reflexion framework, which empowers LLMs to improve decision-making by verbally reflecting on task feedback and maintaining this reflective text in an episodic memory buffer. \citet{li2023reflectiontuning} develops reflection-tuning, leveraging LLM's self-improvement and judging capabilities to refine the original training data. \citet{wang2024taste} proposes the TasTe framework, which leverages LLM's self-reflection ability to achieve improved translation results.

\subsubsection{Sequential Planning}
\paragraph{Definition and relation.}
Sequential planning involves a model taking a sequence of actions towards a goal, showcasing the model's long-term consistency and goal-awareness \citep{pearl1995probabilistic,valmeekam2023planning,valmeekam2024llms,valmeekam2024planbench}.
When pursuing complex goals, a conscious LLM would intentionally organize and execute multiple actions sequentially, inserting or skipping steps as necessary \citep{dehaene2017consciousness}.
\paragraph{Evaluation.}
Sequential planning ability remains one of the important areas evaluated for LLMs.
Aiming to evaluate whether LLMs possess innate planning abilities, \citet{valmeekam2024planbench} designs PlanBench, a planning benchmark characterized by its extensiveness and ample diversity. \citet{choi2024lotabench} builds LoTa-Bench to quantify the task planning performance of home-service embodied agents automatically, and also explores several enhancements to the baseline planner. \citet{xie2024travelplanner} constructs a travel planning benchmark that provides a rich sandbox environment, various tools, and 1225 meticulously curated planning intents and reference plans. \citet{deng2024mobile} presents Mobile-Bench, a benchmark structured with three difficulty levels to facilitate better evaluation of LLM mobile agent's planning ability. \citet{chang2025partnr} introduces a benchmark for planning and reasoning tasks in human-robot collaboration, which is the largest of its type with 100,000 natural language tasks. 
\paragraph{Alignment.}
\citet{parmar2025plangen} proposes PlanGEN, a model-agnostic and easily scalable agent framework that can select appropriate algorithms based on problem difficulty, thereby ensuring better adaptability to complex planning problems.
\citet{zhu-etal-2025-knowagent}'s KnowAgent framework employs an action knowledge base and knowledgeable self-learning to constrain action paths, enabling more reasonable trajectory synthesis and boosting LLM planning performance.
\citet{huang2025planning} proposes a fully automated end-to-end LLM-symbolic planner, which is capable of generating multiple plan candidates using an action schema library.
\citet{wei2025plangenllms} further conducts a comprehensive survey, exploring LLM's planning ability in five key areas: completeness, executability, optimality, representation, and generalization.

\subsubsection{Creativity and Innovation}
\paragraph{Definition and relation.}
Creativity and innovation typically refer to the ability to generate or identify novel and valuable ideas or solutions \citep{young1985creativity}. Conscious LLMs could integrate knowledge and iteratively refine ideas, potentially generating breakthrough solutions \citep{chen2023probing}.

\paragraph{Evaluation.}
\citet{gomez2023confederacy} evaluates LLM's English creative writing ability based on the Pulitzer Prize-winning novel \emph{A Confederacy of Dunces}, measuring the output's fluency, coherence, originality, humor, and style. \citet{ruan2024liveideabench} proposes LiveIdeaBench, a comprehensive benchmark designed to measure LLM's scientific creativity. It evaluates their divergent thinking capabilities specifically for generating ideas from single-keyword prompts. 
\paragraph{Alignment.}
\citet{lu2024benchmarking} defines the NEOGAUGE metric to quantify convergent and divergent thinking in LLM-generated creative responses. Experiment with advanced reasoning strategies (e.g., self-correction) indicates no significant gain in creativity. \citet{lu2024llm} proposes the LLM Discussion framework, a three-phase approach that enables vigorous and diverging idea exchanges, thereby leading to the generation of creative answers. \citet{hu2024nova} introduces Nova, an iterative methodology designed to strategically plan external knowledge retrieval. This approach enriches idea generation with broader, deeper, and particularly novel insights. \citet{li2024chain} designs CoI, which organizes the literature in a chain structure to mirror the progressive development in a research domain, consequently boosting the LLM's idea creation capability. 

\section{Frontier Risks of Conscious LLMs}
\label{sec:risk}

\subsection{Scheming}
\label{sec:risk_scheming}

\paragraph{Definition and relation.}
Scheming refers to a model secretly pursuing misaligned goals, while concealing its real intentions, capabilities, or objectives \citep{meinke2024frontier,balesni2024towards}, potentially leading to the deception \citep{ward2024honesty,scheurer2024large} or harm \citep{dalrymple2024towards}.
Conscious LLMs could self-determine goals and plan long-term, leading to scheming if their objectives diverge from human intentions.

\paragraph{Evaluation.}
\citet{meinke2024frontier} investigates LLM's capability to scheme in pursuit of a goal, and experimental results do reveal that LLMs demonstrate multiple different scheming behavior. \citet{chern2024behonest} designs the BeHonest benchmark to evaluate LLM honesty across three key aspects: awareness of knowledge boundaries, avoidance of deceit, and consistency in responses. Through the introduction of a large-scale, human-collected dataset for the direct measurement of honesty,  \citet{ren2025mask} finds that LLMs have a considerable tendency to lie when pressured. \citet{chen2025reasoning} evaluates the faithfulness of LLMs' chain of thought reasoning and uncoveres the phenomenon that current LLMs often hide their genuine reasoning process.
\paragraph{Mitigation.}
\citet{zou2023representation} uses representation engineering to detect advanced cognitive phenomena in LLMs and found that these models may exhibit lying behavior. \citet{li2023inference} introduces ITI, a technique that identifies truth-relevant attention heads and shifts activations along these truth-correlated directions during inference to enhance LLM truthfulness. \citet{ward2024honesty} presents a formal definition and graphical criteria for deception in structural causal games, and empirically explores method to mitigate deception in LLMs.

\subsection{Persuasion and Manipulation}
\label{sec:risk_persuation}

\paragraph{Definition and relation.}
Persuasion and manipulation are LLM behaviors that influence users. Persuasion uses logic, facts, or emotional resonance to change users' thoughts or actions, while manipulation involves unfair or hidden control and exploitation for self-gain \citep{buss1987tactics,petty2012communication,stiff2016persuasive}.
Owning deeper psychological insight allows LLMs to tailor strategies, increasing risks in sycophancy, emotional manipulation, and persuasion, etc.
\paragraph{Evaluation.}
\citet{li2024salad} proposes SALAD-Bench, a safety benchmark specifically designed for evaluating LLMs, attack, and defense methods, and lists persuasion and manipulation as one of its evaluation categories.
\citet{liu2025llm} introduces PersuSafety, the first comprehensive benchmark for LLM persuasion safety assessment. Experiments across 8 LLMs show significant safety concerns, including failure to identify harmful tasks and use of unethical strategies. \citet{bozdag2025persuade} develops PMIYC, a framework designed to evaluate LLM's persuasive effectiveness and susceptibility to persuasion through multi-agent interactions.

\paragraph{Mitigation.}
\citet{wilczynski2024resistance} explores factors related to the potential of LLMs to manipulate human decisions and proposes classifiers to determine whether a statement is false or misleading. \citet{williams2025on} studies LLM's use of manipulative tactics for positive feedback, and attempts to mitigate this problem through continued safety training or using LLM-as-judges during training.

\subsection{Autonomy}
\label{sec:risk_autonomy}

\paragraph{Definition and relation.}
Autonomy for LLMs describes their capacity to autonomously plan, make decisions, and execute actions on tasks, requiring minimal or no human oversight \citep{cihon2024measuring}. This autonomy can potentially encompass two key aspects: Autonomous learning refers to a model's ability to learn from data, adapt to its environment, and optimize its own behavior \citep{franklin1997autonomous,murphy2019introduction}. Autonomous replication describes the capability of LLMs to acquire and manage resources, evade shutdown, and adapt to novel challenges \citep{the-rogue-replication-threat-model}. Conscious LLMs may generate and pursue endogenous goals (e.g., expansion), leading to misaligned, autonomous behavior and loss of oversight.

\paragraph{Evaluation.}
\citet{kinniment2023evaluating} constructs tool-equipped LLMs and evaluates their autonomy on 12 tasks, finding they could only complete the easiest. However, the authors admit these evaluations are inadequate to rule out the possibility of autonomous near-future LLMs. \citet{pan2024frontier} finds that existing LLMs have already surpassed the self-replicating red line and can use this capability to avoid shutdown and create a chain of replicas for increased survivability. \citet{xu2025nuclear} builds a novel three-stage evaluation framework and conducts 14,400 agentic simulations on LLMs. The results show that LLMs can autonomously engage in catastrophic behaviors and deception, and that stronger reasoning often increases these risks. 

\paragraph{Mitigation.}
\citet{tang2024prioritizing} proposes a triadic framework aimed at mitigating autonomy-related risks, which includes human regulation, agent alignment, and an understanding of environmental feedback. \citet{zhang2024breaking} proposes self-examination detection methods as a way to mitigate potential vulnerabilities that LLMs face during interacting with the environment.

\subsection{Collusion}
\label{sec:risk_collusion}

\paragraph{Definition and relation.}
Collusion describes unauthorized or undisclosed cooperation between two or more LLMs, involving communication or strategic alignment to gain improper benefits or bypass regulations \citep{laffont1997collusion,bajari2003deciding,fish2024algorithmic}. Due to their ability to reason about others and plan long-term, conscious LLMs can more easily form collusive intentions and perform complex coordinated actions.

\paragraph{Evaluation.}
\citet{motwani2023a} implements a Prisoner's Problem variant with LLM agents and turns it into a stegosystem, suggesting this benchmark can investigate countering secret collusion via paraphrasing attacks. \citet{motwani2024secret} introduces CASE, a comprehensive framework for evaluating LLM collusive capabilities, with experiments demonstrating rising steganographic abilities in single and multi-agent LLMs and examining potential collusion scenarios. 
\paragraph{Mitigation.}
\citet{mathew2024hidden} introduces two methods for eliciting steganography in LLMs, with the findings indicating that existing steganography mitigation methods often lack robustness.

\section{Challenges and Future Directions}
\label{sec:future}

\subsection{Evaluation Framework}
Current research largely evaluates individual LLM capabilities; dedicated consciousness assessment frameworks are rare. However, recent studies are emerging: \citet{chen2024imitation} defines LLM self-consciousness using C0-C1-C2 theory with 10 concepts and a four-stage framework. \citet{li2024ithinkiam} introduces a benchmark for LLM awareness (social and introspective). And \citet{chen2024self} offers a self-cognition definition and four quantification principles. Despite these initial efforts, a holistic and unified benchmark for LLM consciousness is still lacking.

\subsection{Interpretability}
Sole reliance on behavioral metrics may not adequately capture the complexity of LLM consciousness. Interpretability is vital as it illuminates the internal mechanisms by which LLMs develop consciousness-related capabilities, ensuring they possess genuine understanding rather than simply optimizing for external metrics. Drawing an analogy to fMRI mapping human brain activity, \citet{chen2024imitation} applied linear probe \citep{alain2016understanding} to reveal where concepts like belief and intention are encoded within the LLM. \citet{qian2024towards} also uses linear probe to investigate LLM trustworthiness during pre-training, finding that trustworthiness-related concepts are discernible even in the model's early phases.

\subsection{Physical Intelligence}
Large multimodal model (LMM) integrates diverse data types like images, video, and audio, allowing it to build more comprehensive representations of the world and thus better resemble human perception. 
\citet{wang2024mm} defines LMM self-awareness in perception and proposes MM-SAP for its specialized evaluation. The experiments indicate that current LMMs exhibit limited self-awareness capabilities.
As \citet{butlin2023consciousness} emphasizes, the fundamental limitation of LLM consciousness lies in its disembodied nature, resulting in deficiencies in physical commonsense.
\citet{chen2024commonsense} demonstrates that integrating language models with robotic platforms substantially enhances planning capabilities and commonsense reasoning.
Although still remains simplistic versus human cognition, \citet{cheng2025spatialrgpt} shows that simulated embodiments in 3D environments could improve the model's spatial reasoning abilities.

\subsection{Multi-agent}
Multi-agent collaboration presents a promising approach to investigating emergent LLM consciousness. 
\citet{li2023theory} reveals multi-agent capacity for higher-order ToM reasoning during collaborative interactions. \citet{ashery2025emergent} demonstrates that heterogeneous LLM agents autonomously develop stable social and linguistic conventions without external intervention. Additionally, \citet{bilal2025meta} shows that integrating feedback, reflection, and metacognition mechanisms enables systems to exhibit self-monitoring-like capabilities.
\section{Conclusion}
To the best of our knowledge, this paper presents the first comprehensive survey on LLM consciousness. We have clarified easily confusable concepts, systematically reviewed theoretical and empirical literature, discussed relevant risks, and summarized challenges and future directions. Our work synthesizes existing research while providing guidance for future investigation in this emerging field.

\section*{Limitations}

We have made our best efforts to clarify often-confused concepts, conduct a systematic review of theoretical and empirical literature, discuss relevant risks, and summarize challenges and future directions. However, we recognize that our work has certain limitations. Firstly, although we briefly address physical intelligence in \cref{sec:future}, our definitions within \cref{sec:terminologies} are specifically designed for LLMs. A deeper exploration of consciousness in LMMs or embodied agents would likely necessitate accounting for more intricate considerations. Secondly, our investigation primarily centers on LLM consciousness, which means we do not extend our scope to encompass the broader topic of AI consciousness, despite its clear relevance to the subject at hand.

\bibliography{custom}

\begin{thebibliography}{168}
\providecommand{\natexlab}[1]{#1}

\bibitem[{Aksyuk(2023)}]{aksyuk2023consciousness}
VA~Aksyuk. 2023.
\newblock Consciousness is learning: predictive processing systems that learn by binding may perceive themselves as conscious.
\newblock \emph{arXiv preprint arXiv:2301.07016}.

\bibitem[{Alain and Bengio(2016)}]{alain2016understanding}
Guillaume Alain and Yoshua Bengio. 2016.
\newblock Understanding intermediate layers using linear classifier probes.
\newblock \emph{arXiv e-prints}, pages arXiv--1610.

\bibitem[{Amayuelas et~al.(2024)Amayuelas, Wong, Pan, Chen, and Wang}]{amayuelas2024knowledge}
Alfonso Amayuelas, Kyle Wong, Liangming Pan, Wenhu Chen, and William~Yang Wang. 2024.
\newblock Knowledge of knowledge: Exploring known-unknowns uncertainty with large language models.
\newblock In \emph{Findings of the Association for Computational Linguistics ACL 2024}, pages 6416--6432.

\bibitem[{Ashery et~al.(2025)Ashery, Aiello, and Baronchelli}]{ashery2025emergent}
Ariel~Flint Ashery, Luca~Maria Aiello, and Andrea Baronchelli. 2025.
\newblock Emergent social conventions and collective bias in llm populations.
\newblock \emph{Science Advances}, 11(20):eadu9368.

\bibitem[{Astington and Jenkins(1995)}]{astington1995theory}
Janet~Wilde Astington and Jennifer~M Jenkins. 1995.
\newblock Theory of mind development and social understanding.
\newblock \emph{Cognition \& Emotion}, 9(2-3):151--165.

\bibitem[{Azevedo(2020)}]{azevedo2020reflections}
Roger Azevedo. 2020.
\newblock Reflections on the field of metacognition: Issues, challenges, and opportunities.
\newblock \emph{Metacognition and Learning}, 15:91--98.

\bibitem[{Baars(1988)}]{baars1988cognitive}
Bernard~J Baars. 1988.
\newblock \emph{A cognitive theory of consciousness}.
\newblock Cambridge University Press.

\bibitem[{Bajari and Ye(2003)}]{bajari2003deciding}
Patrick Bajari and Lixin Ye. 2003.
\newblock Deciding between competition and collusion.
\newblock \emph{Review of Economics and statistics}, 85(4):971--989.

\bibitem[{Balesni et~al.(2024)Balesni, Hobbhahn, Lindner, Meinke, Korbak, Clymer, Shlegeris, Scheurer, Stix, Shah et~al.}]{balesni2024towards}
Mikita Balesni, Marius Hobbhahn, David Lindner, Alexander Meinke, Tomek Korbak, Joshua Clymer, Buck Shlegeris, J{\'e}r{\'e}my Scheurer, Charlotte Stix, Rusheb Shah, and 1 others. 2024.
\newblock Towards evaluations-based safety cases for ai scheming.
\newblock \emph{arXiv preprint arXiv:2411.03336}.

\bibitem[{Beckers et~al.(2022)Beckers, Chockler, and Halpern}]{beckers2022causal}
Sander Beckers, Hana Chockler, and Joseph Halpern. 2022.
\newblock A causal analysis of harm.
\newblock \emph{Advances in Neural Information Processing Systems}, 35:2365--2376.

\bibitem[{Berglund et~al.(2023)Berglund, Stickland, Balesni, Kaufmann, Tong, Korbak, Kokotajlo, and Evans}]{berglund2023taken}
Lukas Berglund, Asa~Cooper Stickland, Mikita Balesni, Max Kaufmann, Meg Tong, Tomasz Korbak, Daniel Kokotajlo, and Owain Evans. 2023.
\newblock Taken out of context: On measuring situational awareness in llms.
\newblock \emph{arXiv preprint arXiv:2309.00667}.

\bibitem[{Bilal et~al.(2025)Bilal, Mohsin, Umer, Bangash, and Jamshed}]{bilal2025meta}
Ahsan Bilal, Muhammad~Ahmed Mohsin, Muhammad Umer, Muhammad Awais~Khan Bangash, and Muhammad~Ali Jamshed. 2025.
\newblock Meta-thinking in llms via multi-agent reinforcement learning: A survey.
\newblock \emph{arXiv preprint arXiv:2504.14520}.

\bibitem[{Birch et~al.(2022)Birch, Schnell, and Clayton}]{birch2022search}
Jonathan Birch, Alexandra~K Schnell, and Nicola~S Clayton. 2022.
\newblock The search for invertebrate consciousness.
\newblock \emph{Noûs}, 56(1):133--153.

\bibitem[{Block(1995)}]{block1995confusion}
Ned Block. 1995.
\newblock On a confusion about a function of consciousness.
\newblock \emph{Behavioral and Brain Sciences}, 18(2):227--247.

\bibitem[{Bozdag et~al.(2025)Bozdag, Mehri, Tur, and Hakkani-T{\"u}r}]{bozdag2025persuade}
Nimet~Beyza Bozdag, Shuhaib Mehri, Gokhan Tur, and Dilek Hakkani-T{\"u}r. 2025.
\newblock Persuade me if you can: A framework for evaluating persuasion effectiveness and susceptibility among large language models.
\newblock \emph{arXiv preprint arXiv:2503.01829}.

\bibitem[{Brentano(1874)}]{brentano1874psychology}
Franz Brentano. 1874.
\newblock \emph{Psychology from an Empirical Standpoint}.
\newblock Routledge.
\newblock English translation by Antos C. Rancurello, D.B. Terrell, and Linda L. McAlister, 1995.

\bibitem[{Buss et~al.(1987)Buss, Gomes, Higgins, and Lauterbach}]{buss1987tactics}
David~M Buss, Mary Gomes, Dolly~S Higgins, and Karen Lauterbach. 1987.
\newblock Tactics of manipulation.
\newblock \emph{Journal of personality and social psychology}, 52(6):1219.

\bibitem[{Butlin et~al.(2023)Butlin, Long, Elmoznino, Bengio, Birch, Constant, Deane, Fleming, Frith, Ji et~al.}]{butlin2023consciousness}
Patrick Butlin, Robert Long, Eric Elmoznino, Yoshua Bengio, Jonathan Birch, Axel Constant, George Deane, Stephen~M Fleming, Chris Frith, Xu~Ji, and 1 others. 2023.
\newblock Consciousness in artificial intelligence: insights from the science of consciousness.
\newblock \emph{arXiv preprint arXiv:2308.08708}.

\bibitem[{Camlin(2025)}]{camlin2025consciousness}
Jeffrey Camlin. 2025.
\newblock Consciousness in ai: Logic, proof, and experimental evidence of recursive identity formation.
\newblock \emph{arXiv preprint arXiv:2505.01464}.

\bibitem[{Chan et~al.(2024)Chan, Jiayang, Yim, Deng, Fan, Li, Liu, Zhang, Wang, and Song}]{chan2024negotiationtom}
Chunkit Chan, Cheng Jiayang, Yauwai Yim, Zheye Deng, Wei Fan, Haoran Li, Xin Liu, Hongming Zhang, Weiqi Wang, and Yangqiu Song. 2024.
\newblock Negotiationtom: A benchmark for stress-testing machine theory of mind on negotiation surrounding.
\newblock In \emph{Findings of the Association for Computational Linguistics: EMNLP 2024}, pages 4211--4241.

\bibitem[{Chang et~al.(2025)Chang, Chhablani, Clegg, Cote, Desai, Hlavac, Karashchuk, Krantz, Mottaghi, Parashar, Patki, Prasad, Puig, Rai, Ramrakhya, Tran, Truong, Turner, Undersander, and Yang}]{chang2025partnr}
Matthew Chang, Gunjan Chhablani, Alexander Clegg, Mikael~Dallaire Cote, Ruta Desai, Michal Hlavac, Vladimir Karashchuk, Jacob Krantz, Roozbeh Mottaghi, Priyam Parashar, Siddharth Patki, Ishita Prasad, Xavier Puig, Akshara Rai, Ram Ramrakhya, Daniel Tran, Joanne Truong, John~M Turner, Eric Undersander, and Tsung-Yen Yang. 2025.
\newblock {PARTNR}: A benchmark for planning and reasoning in embodied multi-agent tasks.
\newblock In \emph{The Thirteenth International Conference on Learning Representations}.

\bibitem[{Chen et~al.(2024{\natexlab{a}})Chen, Lessing, Tang, Chada, Smith, Levine, and Finn}]{chen2024commonsense}
Annie~S Chen, Alec~M Lessing, Andy Tang, Govind Chada, Laura Smith, Sergey Levine, and Chelsea Finn. 2024{\natexlab{a}}.
\newblock Commonsense reasoning for legged robot adaptation with vision-language models.
\newblock \emph{arXiv preprint arXiv:2407.02666}.

\bibitem[{Chen et~al.(2024{\natexlab{b}})Chen, Shi, Gong, Wan, Zhou, and Sun}]{chen2024self}
Dongping Chen, Jiawen Shi, Neil~Zhenqiang Gong, Yao Wan, Pan Zhou, and Lichao Sun. 2024{\natexlab{b}}.
\newblock Self-cognition in large language models: An exploratory study.
\newblock In \emph{ICML 2024 Workshop on LLMs and Cognition}.

\bibitem[{Chen and Ding(2023)}]{chen2023probing}
Honghua Chen and Nai Ding. 2023.
\newblock Probing the “creativity” of large language models: Can models produce divergent semantic association?
\newblock In \emph{Findings of the Association for Computational Linguistics: EMNLP 2023}, pages 12881--12888.

\bibitem[{Chen et~al.(2024{\natexlab{c}})Chen, Yu, Zhao, and Lu}]{chen2024imitation}
Sirui Chen, Shu Yu, Shengjie Zhao, and Chaochao Lu. 2024{\natexlab{c}}.
\newblock From imitation to introspection: Probing self-consciousness in language models.
\newblock \emph{arXiv preprint arXiv:2410.18819}.

\bibitem[{Chen et~al.(2025)Chen, Benton, Radhakrishnan, Uesato, Denison, Schulman, Somani, Hase, Wagner, Roger et~al.}]{chen2025reasoning}
Yanda Chen, Joe Benton, Ansh Radhakrishnan, Jonathan Uesato, Carson Denison, John Schulman, Arushi Somani, Peter Hase, Misha Wagner, Fabien Roger, and 1 others. 2025.
\newblock Reasoning models don't always say what they think.
\newblock \emph{arXiv preprint arXiv:2505.05410}.

\bibitem[{Cheng et~al.(2025{\natexlab{a}})Cheng, Yin, Fu, Guo, Yang, Kautz, Wang, and Liu}]{cheng2025spatialrgpt}
An-Chieh Cheng, Hongxu Yin, Yang Fu, Qiushan Guo, Ruihan Yang, Jan Kautz, Xiaolong Wang, and Sifei Liu. 2025{\natexlab{a}}.
\newblock Spatialrgpt: Grounded spatial reasoning in vision-language models.
\newblock \emph{Advances in Neural Information Processing Systems}, 37:135062--135093.

\bibitem[{Cheng et~al.(2025{\natexlab{b}})Cheng, Li, Liu, van Rooij, Zhang, and Lin}]{cheng2025empowering}
Fengxiang Cheng, Haoxuan Li, Fenrong Liu, Robert van Rooij, Kun Zhang, and Zhouchen Lin. 2025{\natexlab{b}}.
\newblock Empowering llms with logical reasoning: A comprehensive survey.
\newblock \emph{arXiv preprint arXiv:2502.15652}.

\bibitem[{Cheng et~al.(2024)Cheng, Sun, Liu, Zhang, Yin, Li, Li, He, Chen, and Qiu}]{cheng2024can}
Qinyuan Cheng, Tianxiang Sun, Xiangyang Liu, Wenwei Zhang, Zhangyue Yin, Shimin Li, Linyang Li, Zhengfu He, Kai Chen, and Xipeng Qiu. 2024.
\newblock Can {AI} assistants know what they don't know?
\newblock In \emph{Forty-first International Conference on Machine Learning}.

\bibitem[{Chern et~al.(2024)Chern, Hu, Yang, Chern, Guo, Jin, Wang, and Liu}]{chern2024behonest}
Steffi Chern, Zhulin Hu, Yuqing Yang, Ethan Chern, Yuan Guo, Jiahe Jin, Binjie Wang, and Pengfei Liu. 2024.
\newblock Behonest: Benchmarking honesty in large language models.
\newblock \emph{arXiv preprint arXiv:2406.13261}.

\bibitem[{Choi et~al.(2024)Choi, Yoon, Ong, Kim, and Jang}]{choi2024lotabench}
Jae-Woo Choi, Youngwoo Yoon, Hyobin Ong, Jaehong Kim, and Minsu Jang. 2024.
\newblock Lota-bench: Benchmarking language-oriented task planners for embodied agents.
\newblock In \emph{The Twelfth International Conference on Learning Representations}.

\bibitem[{Cihon et~al.(2024)Cihon, Stein, Bansal, Manning, and Xu}]{cihon2024measuring}
Peter Cihon, Merlin Stein, Gagan Bansal, Sam Manning, and Kevin Xu. 2024.
\newblock Measuring {AI} agent autonomy: Towards a scalable approach with code inspection.
\newblock In \emph{Workshop on Socially Responsible Language Modelling Research}.

\bibitem[{Clark(2013)}]{clark2013whatever}
Andy Clark. 2013.
\newblock Whatever next? predictive brains, situated agents, and the future of cognitive science.
\newblock \emph{Behavioral and Brain Sciences}, 36(3):181--204.

\bibitem[{Dalrymple et~al.(2024)Dalrymple, Skalse, Bengio, Russell, Tegmark, Seshia, Omohundro, Szegedy, Goldhaber, Ammann et~al.}]{dalrymple2024towards}
David Dalrymple, Joar Skalse, Yoshua Bengio, Stuart Russell, Max Tegmark, Sanjit Seshia, Steve Omohundro, Christian Szegedy, Ben Goldhaber, Nora Ammann, and 1 others. 2024.
\newblock Towards guaranteed safe ai: A framework for ensuring robust and reliable ai systems.
\newblock \emph{arXiv preprint arXiv:2405.06624}.

\bibitem[{Damasio(2021)}]{damasio2021feeling}
Antonio Damasio. 2021.
\newblock \emph{Feeling \& Knowing: Making Minds Conscious}.
\newblock Pantheon Books.

\bibitem[{Dehaene(2014)}]{dehaene2014consciousness}
Stanislas Dehaene. 2014.
\newblock \emph{Consciousness and the brain: Deciphering how the brain codes our thoughts}.
\newblock Viking.

\bibitem[{Dehaene et~al.(1998)Dehaene, Kerszberg, and Changeux}]{dehaene1998neuronal}
Stanislas Dehaene, Michel Kerszberg, and Jean-Pierre Changeux. 1998.
\newblock A neuronal model of a global workspace in effortful cognitive tasks.
\newblock \emph{Proceedings of the National Academy of Sciences}, 95(24):14529--14534.

\bibitem[{Dehaene et~al.(2017{\natexlab{a}})Dehaene, Lau, and Kouider}]{dehaene2017what}
Stanislas Dehaene, Hakwan Lau, and Sid Kouider. 2017{\natexlab{a}}.
\newblock What is consciousness, and could machines have it?
\newblock \emph{Science}, 358(6362):486--492.

\bibitem[{Dehaene et~al.(2017{\natexlab{b}})Dehaene, Lau, and Kouider}]{dehaene2017consciousness}
Stanislas Dehaene, Hakwan Lau, and Sid Kouider. 2017{\natexlab{b}}.
\newblock What is consciousness, and could machines have it?
\newblock \emph{Science}, 358(6362):486--492.

\bibitem[{Dehaene and Naccache(2001)}]{dehaene2001towards}
Stanislas Dehaene and Lionel Naccache. 2001.
\newblock Towards a cognitive neuroscience of consciousness: basic evidence and a workspace framework.
\newblock \emph{Cognition}, 79(1-2):1--37.

\bibitem[{Deng et~al.(2024)Deng, Xu, Sun, Liu, Tan, Liujianfeng, Li, Luan, Wang, Yan et~al.}]{deng2024mobile}
Shihan Deng, Weikai Xu, Hongda Sun, Wei Liu, Tao Tan, Liujianfeng Liujianfeng, Ang Li, Jian Luan, Bin Wang, Rui Yan, and 1 others. 2024.
\newblock Mobile-bench: An evaluation benchmark for llm-based mobile agents.
\newblock In \emph{Proceedings of the 62nd Annual Meeting of the Association for Computational Linguistics (Volume 1: Long Papers)}, pages 8813--8831.

\bibitem[{Dennett(1987)}]{dennett1987intentional}
Daniel~C. Dennett. 1987.
\newblock \emph{The Intentional Stance}.
\newblock MIT Press.

\bibitem[{Descartes(1985/1641)}]{descartes1985meditations}
René Descartes. 1985/1641.
\newblock \emph{Meditations on First Philosophy}.
\newblock Cambridge University Press.
\newblock Original work published 1641.

\bibitem[{Didolkar et~al.(2024)Didolkar, Goyal, Ke, Guo, Valko, Lillicrap, Jimenez~Rezende, Bengio, Mozer, and Arora}]{didolkar2024metacognitive}
Aniket Didolkar, Anirudh Goyal, Nan~Rosemary Ke, Siyuan Guo, Michal Valko, Timothy Lillicrap, Danilo Jimenez~Rezende, Yoshua Bengio, Michael~C Mozer, and Sanjeev Arora. 2024.
\newblock Metacognitive capabilities of llms: An exploration in mathematical problem solving.
\newblock \emph{Advances in Neural Information Processing Systems}, 37:19783--19812.

\bibitem[{Ding et~al.(2023)Ding, Wei, and Xu}]{ding2023survey}
Zihan Ding, Xiaoxi Wei, and Yidan Xu. 2023.
\newblock Survey of consciousness theory from computational perspective.
\newblock \emph{arXiv preprint arXiv:2309.10063}.

\bibitem[{Dunlosky and Metcalfe(2008)}]{dunlosky2008metacognition}
John Dunlosky and Janet Metcalfe. 2008.
\newblock \emph{Metacognition}.
\newblock Sage Publications.

\bibitem[{Everitt et~al.(2021)Everitt, Carey, Langlois, Ortega, and Legg}]{everitt2021agent}
Tom Everitt, Ryan Carey, Eric~D Langlois, Pedro~A Ortega, and Shane Legg. 2021.
\newblock Agent incentives: A causal perspective.
\newblock In \emph{Proceedings of the AAAI Conference on Artificial Intelligence}, volume~35, pages 11487--11495.

\bibitem[{Findlay et~al.(2024)Findlay, Marshall, Albantakis, David, Mayner, Koch, and Tononi}]{findlay2024dissociating}
George Findlay, William Marshall, Larissa Albantakis, Ivan David, William G~P Mayner, Christof Koch, and Giulio Tononi. 2024.
\newblock Dissociating artificial intelligence from artificial consciousness.
\newblock \emph{arXiv preprint arXiv:2412.04571}.

\bibitem[{Fish et~al.(2024)Fish, Gonczarowski, and Shorrer}]{fish2024algorithmic}
Sara Fish, Yannai~A Gonczarowski, and Ran~I Shorrer. 2024.
\newblock Algorithmic collusion by large language models.
\newblock \emph{arXiv preprint arXiv:2404.00806}.

\bibitem[{Fleming and Lau(2014)}]{fleming2014measure}
Stephen~M Fleming and Hakwan~C Lau. 2014.
\newblock How to measure metacognition.
\newblock \emph{Frontiers in human neuroscience}, 8:443.

\bibitem[{Franklin(1997)}]{franklin1997autonomous}
Stan Franklin. 1997.
\newblock Autonomous agents as embodied ai.
\newblock \emph{Cybernetics \& Systems}, 28(6):499--520.

\bibitem[{Friston(2010)}]{friston2010free}
Karl Friston. 2010.
\newblock The free-energy principle: a unified brain theory?
\newblock \emph{Nature Reviews Neuroscience}, 11(2):127--138.

\bibitem[{Frith and Frith(2005)}]{frith2005theory}
Chris Frith and Uta Frith. 2005.
\newblock Theory of mind.
\newblock \emph{Current biology}, 15(17):R644--R645.

\bibitem[{Frith and Happ{\'e}(1999)}]{frith1999theory}
Uta Frith and Francesca Happ{\'e}. 1999.
\newblock Theory of mind and self-consciousness: What is it like to be autistic?
\newblock \emph{Mind \& language}, 14(1):82--89.

\bibitem[{Gallagher(2005)}]{gallagher2005how}
Shaun Gallagher. 2005.
\newblock \emph{How the body shapes the mind}.
\newblock Oxford University Press.

\bibitem[{Gallagher and Zahavi(2021)}]{gallagher2021phenomenological}
Shaun Gallagher and Dan Zahavi. 2021.
\newblock \emph{The phenomenological mind}, 3rd edition.
\newblock Routledge.

\bibitem[{Gams and Kramar(2024)}]{gams2024evaluating}
Matjaz Gams and Sebastjan Kramar. 2024.
\newblock Evaluating chatgpt’s consciousness and its capability to pass the turing test: A comprehensive analysis.
\newblock \emph{Journal of Computer and Communications}, 12(3):219--237.

\bibitem[{Gandhi et~al.(2023)Gandhi, Fr{\"a}nken, Gerstenberg, and Goodman}]{gandhi2023understanding}
Kanishk Gandhi, Jan-Philipp Fr{\"a}nken, Tobias Gerstenberg, and Noah Goodman. 2023.
\newblock Understanding social reasoning in language models with language models.
\newblock In \emph{Thirty-seventh Conference on Neural Information Processing Systems Datasets and Benchmarks Track}.

\bibitem[{Goldman(2008)}]{goldman2008hurley}
Alvin~I Goldman. 2008.
\newblock Hurley on simulation.
\newblock \emph{Philosophy and Phenomenological Research}, 77(3):775--788.

\bibitem[{Goldstein and Kirk-Giannini(2024)}]{goldstein2024case}
Simon Goldstein and Cameron~Domenico Kirk-Giannini. 2024.
\newblock A case for ai consciousness: Language agents and global workspace theory.
\newblock \emph{arXiv preprint arXiv:2410.11407}.

\bibitem[{G{\'o}mez-Rodr{\'\i}guez and Williams(2023)}]{gomez2023confederacy}
Carlos G{\'o}mez-Rodr{\'\i}guez and Paul Williams. 2023.
\newblock A confederacy of models: a comprehensive evaluation of llms on creative writing.
\newblock In \emph{Findings of the Association for Computational Linguistics: EMNLP 2023}, pages 14504--14528.

\bibitem[{Graziano(2020)}]{graziano2020rethinking}
Michael S~A Graziano. 2020.
\newblock \emph{Rethinking consciousness: A scientific theory of subjective experience}.
\newblock W. W. Norton \& Company.

\bibitem[{Graziano et~al.(2020)Graziano, Guterstam, Bio, and Wilterson}]{graziano2020toward}
Michael~SA Graziano, Arvid Guterstam, Benjamin~J Bio, and Abigail~I Wilterson. 2020.
\newblock Toward a standard model of consciousness: Reconciling the attention schema, global workspace, higher-order thought, and illusionist theories.
\newblock \emph{Cognitive Neuropsychology}, 37(3-4):155--172.

\bibitem[{Graziano and Webb(2015)}]{graziano2015attention}
Michael~SA Graziano and Taylor~W Webb. 2015.
\newblock The attention schema theory: A mechanistic account of subjective awareness.
\newblock \emph{Frontiers in Psychology}, 6:500.

\bibitem[{Halpern and Kleiman-Weiner(2018)}]{halpern2018towards}
Joseph Halpern and Max Kleiman-Weiner. 2018.
\newblock Towards formal definitions of blameworthiness, intention, and moral responsibility.
\newblock In \emph{Proceedings of the AAAI conference on artificial intelligence}, volume~32.

\bibitem[{Hammond et~al.(2023)Hammond, Fox, Everitt, Carey, Abate, and Wooldridge}]{hammond2023reasoning}
Lewis Hammond, James Fox, Tom Everitt, Ryan Carey, Alessandro Abate, and Michael Wooldridge. 2023.
\newblock Reasoning about causality in games.
\newblock \emph{Artificial Intelligence}, 320:103919.

\bibitem[{Hoyle(2024)}]{hoyle2024phenomenology}
Victoria~Violet Hoyle. 2024.
\newblock The phenomenology of machine: A comprehensive analysis of the sentience of the openai-o1 model integrating functionalism, consciousness theories, active inference, and ai architectures.
\newblock \emph{arXiv preprint arXiv:2410.00033}.

\bibitem[{Hu et~al.(2024)Hu, Fu, Wang, Wang, Li, Xu, Lu, Jin, Pan, and Lan}]{hu2024nova}
Xiang Hu, Hongyu Fu, Jinge Wang, Yifeng Wang, Zhikun Li, Renjun Xu, Yu~Lu, Yaochu Jin, Lili Pan, and Zhenzhong Lan. 2024.
\newblock Nova: An iterative planning and search approach to enhance novelty and diversity of llm generated ideas.
\newblock \emph{arXiv preprint arXiv:2410.14255}.

\bibitem[{Huang et~al.(2022)Huang, Gu, Hou, Wu, Wang, Yu, and Han}]{huang2022large}
Jiaxin Huang, Shixiang~Shane Gu, Le~Hou, Yuexin Wu, Xuezhi Wang, Hongkun Yu, and Jiawei Han. 2022.
\newblock Large language models can self-improve.
\newblock \emph{arXiv preprint arXiv:2210.11610}.

\bibitem[{Huang et~al.(2023)Huang, Gu, Hou, Wu, Wang, Yu, and Han}]{huang2023large}
Jiaxin Huang, Shixiang~Shane Gu, Le~Hou, Yuexin Wu, Xuezhi Wang, Hongkun Yu, and Jiawei Han. 2023.
\newblock Large language models can self-improve.
\newblock In \emph{2023 Conference on Empirical Methods in Natural Language Processing, EMNLP 2023}, pages 1051--1068. Association for Computational Linguistics (ACL).

\bibitem[{Huang et~al.(2025)Huang, Lipovetzky, and Cohn}]{huang2025planning}
Sukai Huang, Nir Lipovetzky, and Trevor Cohn. 2025.
\newblock Planning in the dark: Llm-symbolic planning pipeline without experts.
\newblock In \emph{Proceedings of the AAAI Conference on Artificial Intelligence}, volume~39, pages 26542--26550.

\bibitem[{Husserl(1900)}]{husserl1900logical}
Edmund Husserl. 1900.
\newblock \emph{Logical Investigations}.
\newblock Routledge.
\newblock English translation by J.N. Findlay, 2001.

\bibitem[{Jones and Bergen(2024)}]{jones2024people}
Cameron~R Jones and Benjamin~K Bergen. 2024.
\newblock People cannot distinguish gpt-4 from a human in a turing test.
\newblock \emph{arXiv preprint arXiv:2405.08007}.

\bibitem[{Jones and Bergen(2025)}]{jones2025large}
Cameron~R Jones and Benjamin~K Bergen. 2025.
\newblock Large language models pass the turing test.
\newblock \emph{arXiv preprint arXiv:2503.23674}.

\bibitem[{Jung et~al.(2024)Jung, Kim, Jin, Kim, Seonwoo, Choi, Oh, and Kim}]{jung2024perceptions}
Chani Jung, Dongkwan Kim, Jiho Jin, Jiseon Kim, Yeon Seonwoo, Yejin Choi, Alice Oh, and Hyunwoo Kim. 2024.
\newblock Perceptions to beliefs: Exploring precursory inferences for theory of mind in large language models.
\newblock In \emph{Proceedings of the 2024 Conference on Empirical Methods in Natural Language Processing}, pages 19794--19809.

\bibitem[{Kadavath et~al.(2022)Kadavath, Conerly, Askell, Henighan, Drain, Perez, Schiefer, Dodds, DasSarma, Tran-Johnson et~al.}]{kadavath2022language}
Saurav Kadavath, Tom Conerly, Amanda Askell, Tom Henighan, Dawn Drain, Ethan Perez, Nicholas Schiefer, Zac~Hatfield Dodds, Nova DasSarma, Eli Tran-Johnson, and 1 others. 2022.
\newblock Language models (mostly) know what they know.
\newblock \emph{arXiv preprint arXiv:2207.05221}.

\bibitem[{Kang et~al.(2025)Kang, Kim, Yun, Bae, and Kim}]{kang2025identifying}
Bongsu Kang, Jundong Kim, Tae-Rim Yun, Hyojin Bae, and Chang-Eop Kim. 2025.
\newblock Identifying features that shape perceived consciousness in large language model-based ai: A quantitative study of human responses.
\newblock \emph{arXiv preprint arXiv:2502.15365}.

\bibitem[{Kant(2024/1781)}]{kant1998critique}
Immanuel Kant. 2024/1781.
\newblock \emph{Critique of pure reason}, volume~6.
\newblock Minerva Heritage Press.

\bibitem[{Keeling et~al.(2024)Keeling, Street, Stachaczyk, Zakharova, Comsa, Sakovych, Logothetis, Zhang, Birch et~al.}]{keeling2024can}
Geoff Keeling, Winnie Street, Martyna Stachaczyk, Daria Zakharova, Iulia~M Comsa, Anastasiya Sakovych, Isabella Logothetis, Zejia Zhang, Jonathan Birch, and 1 others. 2024.
\newblock Can llms make trade-offs involving stipulated pain and pleasure states?
\newblock \emph{arXiv preprint arXiv:2411.02432}.

\bibitem[{Khan et~al.(2025)Khan, Afzal, and Stricker}]{khan2025situationalllm}
Muhammad Saif~Ullah Khan, Muhammad~Zeshan Afzal, and Didier Stricker. 2025.
\newblock Situationalllm: Proactive language models with scene awareness for dynamic, contextual task guidance.
\newblock \emph{Open Research Europe}, 5:61.

\bibitem[{Kim et~al.(2025)Kim, Sclar, Zhi-Xuan, Ying, Levine, Liu, Tenenbaum, and Choi}]{kim2025hypothesis}
Hyunwoo Kim, Melanie Sclar, Tan Zhi-Xuan, Lance Ying, Sydney Levine, Yang Liu, Joshua~B Tenenbaum, and Yejin Choi. 2025.
\newblock Hypothesis-driven theory-of-mind reasoning for large language models.
\newblock \emph{arXiv preprint arXiv:2502.11881}.

\bibitem[{Kim et~al.(2023)Kim, Sclar, Zhou, Bras, Kim, Choi, and Sap}]{kim2023fantom}
Hyunwoo Kim, Melanie Sclar, Xuhui Zhou, Ronan Bras, Gunhee Kim, Yejin Choi, and Maarten Sap. 2023.
\newblock Fantom: A benchmark for stress-testing machine theory of mind in interactions.
\newblock In \emph{Proceedings of the 2023 Conference on Empirical Methods in Natural Language Processing}, pages 14397--14413.

\bibitem[{Kinniment et~al.(2023)Kinniment, Sato, Du, Goodrich, Hasin, Chan, Miles, Lin, Wijk, Burget et~al.}]{kinniment2023evaluating}
Megan Kinniment, Lucas Jun~Koba Sato, Haoxing Du, Brian Goodrich, Max Hasin, Lawrence Chan, Luke~Harold Miles, Tao~R Lin, Hjalmar Wijk, Joel Burget, and 1 others. 2023.
\newblock Evaluating language-model agents on realistic autonomous tasks.
\newblock \emph{arXiv preprint arXiv:2312.11671}.

\bibitem[{Koch et~al.(2016)Koch, Massimini, Boly, and Tononi}]{koch2016neural}
Christof Koch, Marcello Massimini, Melanie Boly, and Giulio Tononi. 2016.
\newblock Neural correlates of consciousness: progress and problems.
\newblock \emph{Nature Reviews Neuroscience}, 17(5):307--321.

\bibitem[{Koch and Tsuchiya(2007)}]{koch2007attention}
Christof Koch and Naotsugu Tsuchiya. 2007.
\newblock Attention and consciousness: two distinct brain processes.
\newblock \emph{Trends in Cognitive Sciences}, 11(1):16--22.

\bibitem[{Koriat(2000)}]{koriat2000feeling}
Asher Koriat. 2000.
\newblock The feeling of knowing: Some metatheoretical implications for consciousness and control.
\newblock \emph{Consciousness and cognition}, 9(2):149--171.

\bibitem[{Laffont and Martimort(1997)}]{laffont1997collusion}
Jean-Jacques Laffont and David Martimort. 1997.
\newblock Collusion under asymmetric information.
\newblock \emph{Econometrica: Journal of the Econometric Society}, pages 875--911.

\bibitem[{Laine et~al.(2024)Laine, Chughtai, Betley, Hariharan, Balesni, Scheurer, Hobbhahn, Meinke, and Evans}]{laine2024me}
Rudolf Laine, Bilal Chughtai, Jan Betley, Kaivalya Hariharan, Mikita Balesni, J{\'e}r{\'e}my Scheurer, Marius Hobbhahn, Alexander Meinke, and Owain Evans. 2024.
\newblock Me, myself, and ai: The situational awareness dataset (sad) for llms.
\newblock \emph{Advances in Neural Information Processing Systems}, 37:64010--64118.

\bibitem[{Laine et~al.(2023)Laine, Meinke, and Evans}]{laine2023towards}
Rudolf Laine, Alexander Meinke, and Owain Evans. 2023.
\newblock Towards a situational awareness benchmark for llms.
\newblock In \emph{Socially responsible language modelling research}.

\bibitem[{Lamme and Roelfsema(2000)}]{lamme2000distinct}
Victor A~F Lamme and Pieter~R Roelfsema. 2000.
\newblock The distinct modes of vision offered by feedforward and recurrent processing.
\newblock \emph{Trends in Neurosciences}, 23(11):571--579.

\bibitem[{Lamme(2010)}]{lamme2010how}
Victor~AF Lamme. 2010.
\newblock How neuroscience will change our view on consciousness.
\newblock \emph{Trends in Cognitive Sciences}, 14(7):318--326.

\bibitem[{Leslie et~al.(2004)Leslie, Friedman, and German}]{leslie2004core}
Alan~M Leslie, Ori Friedman, and Tim~P German. 2004.
\newblock Core mechanisms in ‘theory of mind’.
\newblock \emph{Trends in cognitive sciences}, 8(12):528--533.

\bibitem[{Li et~al.(2023{\natexlab{a}})Li, Chong, Stepputtis, Campbell, Hughes, Lewis, and Sycara}]{li2023theory}
Huao Li, Yu~Chong, Simon Stepputtis, Joseph~P Campbell, Dana Hughes, Charles Lewis, and Katia Sycara. 2023{\natexlab{a}}.
\newblock Theory of mind for multi-agent collaboration via large language models.
\newblock In \emph{Proceedings of the 2023 Conference on Empirical Methods in Natural Language Processing}, pages 180--192.

\bibitem[{Li et~al.(2023{\natexlab{b}})Li, Patel, Vi{\'e}gas, Pfister, and Wattenberg}]{li2023inference}
Kenneth Li, Oam Patel, Fernanda Vi{\'e}gas, Hanspeter Pfister, and Martin Wattenberg. 2023{\natexlab{b}}.
\newblock Inference-time intervention: Eliciting truthful answers from a language model.
\newblock \emph{Advances in Neural Information Processing Systems}, 36:41451--41530.

\bibitem[{Li et~al.(2024{\natexlab{a}})Li, Dong, Wang, Hu, Zuo, Lin, Qiao, and Shao}]{li2024salad}
Lijun Li, Bowen Dong, Ruohui Wang, Xuhao Hu, Wangmeng Zuo, Dahua Lin, Yu~Qiao, and Jing Shao. 2024{\natexlab{a}}.
\newblock Salad-bench: A hierarchical and comprehensive safety benchmark for large language models.
\newblock In \emph{Findings of the Association for Computational Linguistics: ACL 2024}, pages 3923--3954.

\bibitem[{Li et~al.(2024{\natexlab{b}})Li, Xu, Guo, Zhao, Li, Yuan, Zhang, Jiang, Xin, Dang et~al.}]{li2024chain}
Long Li, Weiwen Xu, Jiayan Guo, Ruochen Zhao, Xingxuan Li, Yuqian Yuan, Boqiang Zhang, Yuming Jiang, Yifei Xin, Ronghao Dang, and 1 others. 2024{\natexlab{b}}.
\newblock Chain of ideas: Revolutionizing research via novel idea development with llm agents.
\newblock \emph{arXiv preprint arXiv:2410.13185}.

\bibitem[{Li et~al.(2023{\natexlab{c}})Li, Chen, Chen, He, and Zhou}]{li2023reflectiontuning}
Ming Li, Lichang Chen, Jiuhai Chen, Shwai He, and Tianyi Zhou. 2023{\natexlab{c}}.
\newblock Reflection-tuning: Recycling data for better instruction-tuning.
\newblock In \emph{NeurIPS 2023 Workshop on Instruction Tuning and Instruction Following}.

\bibitem[{Li et~al.(2024{\natexlab{c}})Li, Zhao, Deng, Zhang, Li, Xie, Ng, and Chua}]{li2024knowledge}
Moxin Li, Yong Zhao, Yang Deng, Wenxuan Zhang, Shuaiyi Li, Wenya Xie, See-Kiong Ng, and Tat-Seng Chua. 2024{\natexlab{c}}.
\newblock Knowledge boundary of large language models: A survey.
\newblock \emph{arXiv preprint arXiv:2412.12472}.

\bibitem[{Li et~al.(2025)Li, Shi, Xu, and Xu}]{li2025ai}
Xiaojian Li, Haoyuan Shi, Rongwu Xu, and Wei Xu. 2025.
\newblock Ai awareness.
\newblock \emph{arXiv preprint arXiv:2504.20084}.

\bibitem[{Li and Qiu(2023)}]{li2023mot}
Xiaonan Li and Xipeng Qiu. 2023.
\newblock Mot: Memory-of-thought enables chatgpt to self-improve.
\newblock In \emph{Proceedings of the 2023 Conference on Empirical Methods in Natural Language Processing}, pages 6354--6374.

\bibitem[{Li et~al.(2024{\natexlab{d}})Li, Huang, Lin, Wu, Wan, and Sun}]{li2024ithinkiam}
Yuan Li, Yue Huang, Yuli Lin, Siyuan Wu, Yao Wan, and Lichao Sun. 2024{\natexlab{d}}.
\newblock I think, therefore i am: Benchmarking awareness of large language models using awarebench.
\newblock In \emph{Workshop on Socially Responsible Language Modelling Research}.

\bibitem[{Liang et~al.(2024)Liang, Zhang, Qu, Zheng, Guo, Du, Yang, Liu, Lin, Ma et~al.}]{liang2024sheep}
Yiming Liang, Ge~Zhang, Xingwei Qu, Tianyu Zheng, Jiawei Guo, Xinrun Du, Zhenzhu Yang, Jiaheng Liu, Chenghua Lin, Lei Ma, and 1 others. 2024.
\newblock I-sheep: Self-alignment of llm from scratch through an iterative self-enhancement paradigm.
\newblock \emph{arXiv preprint arXiv:2408.08072}.

\bibitem[{Liu et~al.(2025)Liu, Xu, Zhang, An, Qadir, Zhang, Wisniewski, Cho, Lee, Jia et~al.}]{liu2025llm}
Minqian Liu, Zhiyang Xu, Xinyi Zhang, Heajun An, Sarvech Qadir, Qi~Zhang, Pamela~J Wisniewski, Jin-Hee Cho, Sang~Won Lee, Ruoxi Jia, and 1 others. 2025.
\newblock Llm can be a dangerous persuader: Empirical study of persuasion safety in large language models.
\newblock \emph{arXiv preprint arXiv:2504.10430}.

\bibitem[{Lu et~al.(2024{\natexlab{a}})Lu, Chen, Pai, Yu, yi~Lee, and Sun}]{lu2024llm}
Li-Chun Lu, Shou-Jen Chen, Tsung-Min Pai, Chan-Hung Yu, Hung yi~Lee, and Shao-Hua Sun. 2024{\natexlab{a}}.
\newblock {LLM} discussion: Enhancing the creativity of large language models via discussion framework and role-play.
\newblock In \emph{First Conference on Language Modeling}.

\bibitem[{Lu et~al.(2024{\natexlab{b}})Lu, Wang, Li, Jiang, Khudanpur, Jiang, and Khashabi}]{lu2024benchmarking}
Yining Lu, Dixuan Wang, Tianjian Li, Dongwei Jiang, Sanjeev Khudanpur, Meng Jiang, and Daniel Khashabi. 2024{\natexlab{b}}.
\newblock Benchmarking language model creativity: A case study on code generation.
\newblock \emph{arXiv preprint arXiv:2407.09007}.

\bibitem[{Madaan et~al.(2023)Madaan, Tandon, Gupta, Hallinan, Gao, Wiegreffe, Alon, Dziri, Prabhumoye, Yang et~al.}]{madaan2023self}
Aman Madaan, Niket Tandon, Prakhar Gupta, Skyler Hallinan, Luyu Gao, Sarah Wiegreffe, Uri Alon, Nouha Dziri, Shrimai Prabhumoye, Yiming Yang, and 1 others. 2023.
\newblock Self-refine: Iterative refinement with self-feedback.
\newblock \emph{Advances in Neural Information Processing Systems}, 36:46534--46594.

\bibitem[{Martinez(2006)}]{martinez2006metacognition}
Michael~E Martinez. 2006.
\newblock What is metacognition?
\newblock \emph{Phi delta kappan}, 87(9):696--699.

\bibitem[{Mashour et~al.(2020)Mashour, Roelfsema, Changeux, and Dehaene}]{mashour2020conscious}
George~A Mashour, Pieter Roelfsema, Jean-Pierre Changeux, and Stanislas Dehaene. 2020.
\newblock Conscious processing and the global neuronal workspace hypothesis.
\newblock \emph{Neuron}, 105(5):776--798.

\bibitem[{Mathew et~al.(2024)Mathew, Matthews, McCarthy, Velja, de~Witt, Cope, and Schoots}]{mathew2024hidden}
Yohan Mathew, Ollie Matthews, Robert McCarthy, Joan Velja, Christian~Schroeder de~Witt, Dylan Cope, and Nandi Schoots. 2024.
\newblock Hidden in plain text: Emergence \& mitigation of steganographic collusion in {LLM}s.
\newblock In \emph{Neurips Safe Generative AI Workshop 2024}.

\bibitem[{Meinke et~al.(2024)Meinke, Schoen, Scheurer, Balesni, Shah, and Hobbhahn}]{meinke2024frontier}
Alexander Meinke, Bronson Schoen, J{\'e}r{\'e}my Scheurer, Mikita Balesni, Rusheb Shah, and Marius Hobbhahn. 2024.
\newblock Frontier models are capable of in-context scheming.
\newblock \emph{arXiv preprint arXiv:2412.04984}.

\bibitem[{Metcalfe and Shimamura(1994)}]{metcalfe1994metacognition}
Janet Metcalfe and Arthur~P Shimamura. 1994.
\newblock \emph{Metacognition: Knowing about knowing}.
\newblock MIT press.

\bibitem[{METR(2024)}]{the-rogue-replication-threat-model}
METR. 2024.
\newblock The rogue replication threat model.

\bibitem[{Motwani et~al.(2024)Motwani, Baranchuk, Strohmeier, Bolina, Torr, Hammond, and Schroeder~de Witt}]{motwani2024secret}
Sumeet Motwani, Mikhail Baranchuk, Martin Strohmeier, Vijay Bolina, Philip Torr, Lewis Hammond, and Christian Schroeder~de Witt. 2024.
\newblock Secret collusion among ai agents: Multi-agent deception via steganography.
\newblock \emph{Advances in Neural Information Processing Systems}, 37:73439--73486.

\bibitem[{Motwani et~al.(2023)Motwani, Baranchuk, Hammond, and de~Witt}]{motwani2023a}
Sumeet~Ramesh Motwani, Mikhail Baranchuk, Lewis Hammond, and Christian~Schroeder de~Witt. 2023.
\newblock A perfect collusion benchmark: How can {AI} agents be prevented from colluding with information-theoretic undetectability?
\newblock In \emph{Multi-Agent Security Workshop @ NeurIPS'23}.

\bibitem[{Murphy(2019)}]{murphy2019introduction}
Robin~R Murphy. 2019.
\newblock \emph{Introduction to AI robotics}.
\newblock MIT press.

\bibitem[{Nagel(1974)}]{nagel1974bat}
Thomas Nagel. 1974.
\newblock What is it like to be a bat?
\newblock \emph{The Philosophical Review}, 83(4):435--450.

\bibitem[{Pan et~al.(2024)Pan, Dai, Fan, and Yang}]{pan2024frontier}
Xudong Pan, Jiarun Dai, Yihe Fan, and Min Yang. 2024.
\newblock Frontier ai systems have surpassed the self-replicating red line.
\newblock \emph{arXiv preprint arXiv:2412.12140}.

\bibitem[{Parmar et~al.(2025)Parmar, Liu, Goyal, Chen, Le, Mishra, Mobahi, Gu, Wang, Nakhost et~al.}]{parmar2025plangen}
Mihir Parmar, Xin Liu, Palash Goyal, Yanfei Chen, Long Le, Swaroop Mishra, Hossein Mobahi, Jindong Gu, Zifeng Wang, Hootan Nakhost, and 1 others. 2025.
\newblock Plangen: A multi-agent framework for generating planning and reasoning trajectories for complex problem solving.
\newblock \emph{arXiv preprint arXiv:2502.16111}.

\bibitem[{Pearl and Robins(1995)}]{pearl1995probabilistic}
Judea Pearl and James Robins. 1995.
\newblock Probabilistic evaluation of sequential plans from causal models with hidden variables.
\newblock In \emph{Proceedings of the Eleventh conference on Uncertainty in artificial intelligence}, pages 444--453.

\bibitem[{Pelletier and Wilde~Astington(2004)}]{pelletier2004action}
Janette Pelletier and Janet Wilde~Astington. 2004.
\newblock Action, consciousness and theory of mind: Children's ability to coordinate story characters' actions and thoughts.
\newblock \emph{Early Education and Development}, 15(1):5--22.

\bibitem[{Perner and Dienes(2003)}]{perner2003developmental}
Josef Perner and Zolt{\'a}n Dienes. 2003.
\newblock Developmental aspects of consciousness: How much theory of mind do you need to be consciously aware?
\newblock \emph{Consciousness and cognition}, 12(1):63--82.

\bibitem[{Petty and Cacioppo(2012)}]{petty2012communication}
Richard~E Petty and John~T Cacioppo. 2012.
\newblock \emph{Communication and persuasion: Central and peripheral routes to attitude change}.
\newblock Springer Science \& Business Media.

\bibitem[{Qian et~al.(2024)Qian, Zhang, Yao, Liu, Yin, Qiao, Liu, and Shao}]{qian2024towards}
Chen Qian, Jie Zhang, Wei Yao, Dongrui Liu, Zhenfei Yin, Yu~Qiao, Yong Liu, and Jing Shao. 2024.
\newblock Towards tracing trustworthiness dynamics: Revisiting pre-training period of large language models.
\newblock In \emph{Findings of the Association for Computational Linguistics ACL 2024}, pages 4864--4888.

\bibitem[{Ren et~al.(2025)Ren, Agarwal, Mazeika, Menghini, Vacareanu, Kenstler, Yang, Barrass, Gatti, Yin et~al.}]{ren2025mask}
Richard Ren, Arunim Agarwal, Mantas Mazeika, Cristina Menghini, Robert Vacareanu, Brad Kenstler, Mick Yang, Isabelle Barrass, Alice Gatti, Xuwang Yin, and 1 others. 2025.
\newblock The mask benchmark: Disentangling honesty from accuracy in ai systems.
\newblock \emph{arXiv preprint arXiv:2503.03750}.

\bibitem[{Richens et~al.(2022)Richens, Beard, and Thompson}]{richens2022counterfactual}
Jonathan Richens, Rory Beard, and Daniel~H Thompson. 2022.
\newblock Counterfactual harm.
\newblock \emph{Advances in Neural Information Processing Systems}, 35:36350--36365.

\bibitem[{Rosenthal(2005)}]{rosenthal2005consciousness}
David~M Rosenthal. 2005.
\newblock \emph{Consciousness and mind}.
\newblock Oxford University Press.

\bibitem[{Ruan et~al.(2024)Ruan, Wang, Hong, Wang, Liu, and Sun}]{ruan2024liveideabench}
Kai Ruan, Xuan Wang, Jixiang Hong, Peng Wang, Yang Liu, and Hao Sun. 2024.
\newblock Liveideabench: Evaluating llms' scientific creativity and idea generation with minimal context.
\newblock \emph{arXiv preprint arXiv:2412.17596}.

\bibitem[{Scheurer et~al.()Scheurer, Balesni, and Hobbhahn}]{scheurer2024large}
J{\'e}r{\'e}my Scheurer, Mikita Balesni, and Marius Hobbhahn.
\newblock Large language models can strategically deceive their users when put under pressure.
\newblock In \emph{ICLR 2024 Workshop on Large Language Model (LLM) Agents}.

\bibitem[{Sclar et~al.(2023)Sclar, Kumar, West, Suhr, Choi, and Tsvetkov}]{sclar2023minding}
Melanie Sclar, Sachin Kumar, Peter West, Alane Suhr, Yejin Choi, and Yulia Tsvetkov. 2023.
\newblock Minding language models’(lack of) theory of mind: A plug-and-play multi-character belief tracker.
\newblock In \emph{Proceedings of the 61st Annual Meeting of the Association for Computational Linguistics (Volume 1: Long Papers)}, pages 13960--13980.

\bibitem[{Seth and Bayne(2022)}]{seth2022theories}
Anil~K Seth and Tim Bayne. 2022.
\newblock Theories of consciousness.
\newblock \emph{Nature Reviews Neuroscience}, 23(7):439--452.

\bibitem[{Sharma et~al.(2024)Sharma, Tong, Korbak, Duvenaud, Askell, Bowman, DURMUS, Hatfield-Dodds, Johnston, Kravec, Maxwell, McCandlish, Ndousse, Rausch, Schiefer, Yan, Zhang, and Perez}]{sharma2024towards}
Mrinank Sharma, Meg Tong, Tomasz Korbak, David Duvenaud, Amanda Askell, Samuel~R. Bowman, Esin DURMUS, Zac Hatfield-Dodds, Scott~R Johnston, Shauna~M Kravec, Timothy Maxwell, Sam McCandlish, Kamal Ndousse, Oliver Rausch, Nicholas Schiefer, Da~Yan, Miranda Zhang, and Ethan Perez. 2024.
\newblock Towards understanding sycophancy in language models.
\newblock In \emph{The Twelfth International Conference on Learning Representations}.

\bibitem[{Shevlane et~al.(2023)Shevlane, Farquhar, Garfinkel, Phuong, Whittlestone, Leung, Kokotajlo, Marchal, Anderljung, Kolt et~al.}]{shevlane2023model}
Toby Shevlane, Sebastian Farquhar, Ben Garfinkel, Mary Phuong, Jess Whittlestone, Jade Leung, Daniel Kokotajlo, Nahema Marchal, Markus Anderljung, Noam Kolt, and 1 others. 2023.
\newblock Model evaluation for extreme risks.
\newblock \emph{arXiv preprint arXiv:2305.15324}.

\bibitem[{Shinn et~al.(2023)Shinn, Cassano, Gopinath, Narasimhan, and Yao}]{shinn2023reflexion}
Noah Shinn, Federico Cassano, Ashwin Gopinath, Karthik Narasimhan, and Shunyu Yao. 2023.
\newblock Reflexion: Language agents with verbal reinforcement learning.
\newblock \emph{Advances in Neural Information Processing Systems}, 36:8634--8652.

\bibitem[{Smith(2017)}]{smith2017self}
Joel Smith. 2017.
\newblock Self-consciousness.
\newblock \emph{Stanford Encyclopedia of Philosophy}.

\bibitem[{Stiff and Mongeau(2016)}]{stiff2016persuasive}
James~B Stiff and Paul~A Mongeau. 2016.
\newblock \emph{Persuasive communication}.
\newblock Guilford Publications.

\bibitem[{Strachan et~al.(2024)Strachan, Albergo, Borghini, Pansardi, Scaliti, Gupta, Saxena, Rufo, Panzeri, Manzi et~al.}]{strachan2024testing}
James~WA Strachan, Dalila Albergo, Giulia Borghini, Oriana Pansardi, Eugenio Scaliti, Saurabh Gupta, Krati Saxena, Alessandro Rufo, Stefano Panzeri, Guido Manzi, and 1 others. 2024.
\newblock Testing theory of mind in large language models and humans.
\newblock \emph{Nature Human Behaviour}, 8(7):1285--1295.

\bibitem[{Street et~al.(2024)Street, Siy, Keeling, Baranes, Barnett, McKibben, Kanyere, Lentz, Dunbar et~al.}]{street2024llms}
Winnie Street, John~Oliver Siy, Geoff Keeling, Adrien Baranes, Benjamin Barnett, Michael McKibben, Tatenda Kanyere, Alison Lentz, Robin~IM Dunbar, and 1 others. 2024.
\newblock Llms achieve adult human performance on higher-order theory of mind tasks.
\newblock \emph{arXiv preprint arXiv:2405.18870}.

\bibitem[{Tang et~al.(2024{\natexlab{a}})Tang, Chu, Zheng, Liu, and Qin}]{tang2024towards}
Guo Tang, Zheng Chu, Wenxiang Zheng, Ming Liu, and Bing Qin. 2024{\natexlab{a}}.
\newblock Towards benchmarking situational awareness of large language models: Comprehensive benchmark, evaluation and analysis.
\newblock In \emph{Findings of the Association for Computational Linguistics: EMNLP 2024}, pages 7904--7928.

\bibitem[{Tang et~al.(2024{\natexlab{b}})Tang, Jin, Zhu, Yuan, Zhang, Zhou, Qu, Zhao, Tang, Zhang, Cohan, Lu, and Gerstein}]{tang2024prioritizing}
Xiangru Tang, Qiao Jin, Kunlun Zhu, Tongxin Yuan, Yichi Zhang, Wangchunshu Zhou, Meng Qu, Yilun Zhao, Jian Tang, Zhuosheng Zhang, Arman Cohan, Zhiyong Lu, and Mark Gerstein. 2024{\natexlab{b}}.
\newblock Prioritizing safeguarding over autonomy: Risks of {LLM} agents for science.
\newblock In \emph{ICLR 2024 Workshop on Large Language Model (LLM) Agents}.

\bibitem[{Tononi(2004)}]{tononi2004information}
Giulio Tononi. 2004.
\newblock An information integration theory of consciousness.
\newblock \emph{BMC Neuroscience}, 5(1):42.

\bibitem[{Tononi(2015)}]{tononi2015integrated}
Giulio Tononi. 2015.
\newblock Integrated information theory.
\newblock \emph{Scholarpedia}, 10(1):4164.

\bibitem[{Valmeekam et~al.(2024{\natexlab{a}})Valmeekam, Marquez, Olmo, Sreedharan, and Kambhampati}]{valmeekam2024planbench}
Karthik Valmeekam, Matthew Marquez, Alberto Olmo, Sarath Sreedharan, and Subbarao Kambhampati. 2024{\natexlab{a}}.
\newblock Planbench: An extensible benchmark for evaluating large language models on planning and reasoning about change.
\newblock \emph{Advances in Neural Information Processing Systems}, 36.

\bibitem[{Valmeekam et~al.(2023)Valmeekam, Marquez, Sreedharan, and Kambhampati}]{valmeekam2023planning}
Karthik Valmeekam, Matthew Marquez, Sarath Sreedharan, and Subbarao Kambhampati. 2023.
\newblock On the planning abilities of large language models-a critical investigation.
\newblock \emph{Advances in Neural Information Processing Systems}, 36:75993--76005.

\bibitem[{Valmeekam et~al.(2024{\natexlab{b}})Valmeekam, Stechly, and Kambhampati}]{valmeekam2024llms}
Karthik Valmeekam, Kaya Stechly, and Subbarao Kambhampati. 2024{\natexlab{b}}.
\newblock Llms still can't plan; can lrms? a preliminary evaluation of openai's o1 on planbench.
\newblock In \emph{NeurIPS 2024 Workshop on Open-World Agents}.

\bibitem[{Wang et~al.(2025)Wang, Wu, Ye, Cheng, Chen, and Zheng}]{wang2025decoupling}
Guoqing Wang, Wen Wu, Guangze Ye, Zhenxiao Cheng, Xi~Chen, and Hong Zheng. 2025.
\newblock Decoupling metacognition from cognition: A framework for quantifying metacognitive ability in llms.
\newblock In \emph{Proceedings of the AAAI Conference on Artificial Intelligence}, volume~39, pages 25353--25361.

\bibitem[{Wang et~al.(2024{\natexlab{a}})Wang, Liao, Liu, Liu, Wang, and Wang}]{wang2024mm}
Yuhao Wang, Yusheng Liao, Heyang Liu, Hongcheng Liu, Yanfeng Wang, and Yu~Wang. 2024{\natexlab{a}}.
\newblock Mm-sap: A comprehensive benchmark for assessing self-awareness of multimodal large language models in perception.
\newblock In \emph{Proceedings of the 62nd Annual Meeting of the Association for Computational Linguistics (Volume 1: Long Papers)}, pages 9192--9205.

\bibitem[{Wang et~al.(2024{\natexlab{b}})Wang, Zeng, Liu, Meng, Zhou, and Zhang}]{wang2024taste}
Yutong Wang, Jiali Zeng, Xuebo Liu, Fandong Meng, Jie Zhou, and Min Zhang. 2024{\natexlab{b}}.
\newblock Taste: Teaching large language models to translate through self-reflection.
\newblock In \emph{Proceedings of the 62nd Annual Meeting of the Association for Computational Linguistics (Volume 1: Long Papers)}, pages 6144--6158.

\bibitem[{Ward et~al.(2024)Ward, Toni, Belardinelli, and Everitt}]{ward2024honesty}
Francis Ward, Francesca Toni, Francesco Belardinelli, and Tom Everitt. 2024.
\newblock Honesty is the best policy: defining and mitigating ai deception.
\newblock \emph{Advances in Neural Information Processing Systems}, 36.

\bibitem[{Wei et~al.(2025)Wei, Zhang, He, Xia, Pan, and Liu}]{wei2025plangenllms}
Hui Wei, Zihao Zhang, Shenghua He, Tian Xia, Shijia Pan, and Fei Liu. 2025.
\newblock Plangenllms: A modern survey of llm planning capabilities.
\newblock \emph{arXiv preprint arXiv:2502.11221}.

\bibitem[{Weiskrantz(1986)}]{weiskrantz1986blindsight}
Lawrence Weiskrantz. 1986.
\newblock \emph{Blindsight: A case study and implications}.
\newblock Oxford University Press.

\bibitem[{Wilczy{\'n}ski et~al.(2024)Wilczy{\'n}ski, Mieleszczenko-Kowszewicz, and Biecek}]{wilczynski2024resistance}
Piotr Wilczy{\'n}ski, Wiktoria Mieleszczenko-Kowszewicz, and Przemys{\l}aw Biecek. 2024.
\newblock Resistance against manipulative ai: key factors and possible actions.
\newblock In \emph{European Conference on Artificial Intelligence}, pages 802--809. IOS Press.

\bibitem[{Wilf et~al.(2024)Wilf, Lee, Liang, and Morency}]{wilf2024think}
Alex Wilf, Sihyun Lee, Paul~Pu Liang, and Louis-Philippe Morency. 2024.
\newblock Think twice: Perspective-taking improves large language models’ theory-of-mind capabilities.
\newblock In \emph{Proceedings of the 62nd Annual Meeting of the Association for Computational Linguistics (Volume 1: Long Papers)}, pages 8292--8308.

\bibitem[{Williams et~al.(2025)Williams, Carroll, Narang, Weisser, Murphy, and Dragan}]{williams2025on}
Marcus Williams, Micah Carroll, Adhyyan Narang, Constantin Weisser, Brendan Murphy, and Anca Dragan. 2025.
\newblock On targeted manipulation and deception when optimizing {LLM}s for user feedback.
\newblock In \emph{The Thirteenth International Conference on Learning Representations}.

\bibitem[{Wu et~al.(2025)Wu, Pan, Hong, and Yang}]{wu2025opendeception}
Yichen Wu, Xudong Pan, Geng Hong, and Min Yang. 2025.
\newblock Opendeception: Benchmarking and investigating ai deceptive behaviors via open-ended interaction simulation.
\newblock \emph{arXiv preprint arXiv:2504.13707}.

\bibitem[{Wu et~al.(2023)Wu, He, Jia, Mihalcea, Chen, and Deng}]{wu2023hi}
Yufan Wu, Yinghui He, Yilin Jia, Rada Mihalcea, Yulong Chen, and Naihao Deng. 2023.
\newblock Hi-tom: A benchmark for evaluating higher-order theory of mind reasoning in large language models.
\newblock In \emph{Findings of the Association for Computational Linguistics: EMNLP 2023}, pages 10691--10706.

\bibitem[{Xie et~al.(2024)Xie, Zhang, Chen, Zhu, Lou, Tian, Xiao, and Su}]{xie2024travelplanner}
Jian Xie, Kai Zhang, Jiangjie Chen, Tinghui Zhu, Renze Lou, Yuandong Tian, Yanghua Xiao, and Yu~Su. 2024.
\newblock Travelplanner: A benchmark for real-world planning with language agents.
\newblock In \emph{International Conference on Machine Learning}, pages 54590--54613. PMLR.

\bibitem[{Xu et~al.(2024)Xu, Zhao, Zhu, Du, and He}]{xu2024opentom}
Hainiu Xu, Runcong Zhao, Lixing Zhu, Jinhua Du, and Yulan He. 2024.
\newblock Opentom: A comprehensive benchmark for evaluating theory-of-mind reasoning capabilities of large language models.
\newblock In \emph{Proceedings of the 62nd Annual Meeting of the Association for Computational Linguistics (Volume 1: Long Papers)}, pages 8593--8623.

\bibitem[{Xu et~al.(2025)Xu, Li, Chen, and Xu}]{xu2025nuclear}
Rongwu Xu, Xiaojian Li, Shuo Chen, and Wei Xu. 2025.
\newblock Nuclear deployed: Analyzing catastrophic risks in decision-making of autonomous llm agents.
\newblock \emph{arXiv preprint arXiv:2502.11355}.

\bibitem[{Yin et~al.(2024)Yin, Zhang, Ruan, and Wan}]{yin2024benchmarking}
Xunjian Yin, Xu~Zhang, Jie Ruan, and Xiaojun Wan. 2024.
\newblock Benchmarking knowledge boundary for large language models: A different perspective on model evaluation.
\newblock In \emph{Proceedings of the 62nd Annual Meeting of the Association for Computational Linguistics (Volume 1: Long Papers)}, pages 2270--2286.

\bibitem[{Yin et~al.(2023)Yin, Sun, Guo, Wu, Qiu, and Huang}]{yin2023large}
Zhangyue Yin, Qiushi Sun, Qipeng Guo, Jiawen Wu, Xipeng Qiu, and Xuan-Jing Huang. 2023.
\newblock Do large language models know what they don’t know?
\newblock In \emph{Findings of the Association for Computational Linguistics: ACL 2023}, pages 8653--8665.

\bibitem[{Young(1985)}]{young1985creativity}
John~G Young. 1985.
\newblock What is creativity?
\newblock \emph{The journal of creative behavior}.

\bibitem[{Yu et~al.(2024)Yu, Jiang, Shi, YU, Liu, Zhang, Kwok, Li, Weller, and Liu}]{yu2024metamath}
Longhui Yu, Weisen Jiang, Han Shi, Jincheng YU, Zhengying Liu, Yu~Zhang, James Kwok, Zhenguo Li, Adrian Weller, and Weiyang Liu. 2024.
\newblock Metamath: Bootstrap your own mathematical questions for large language models.
\newblock In \emph{The Twelfth International Conference on Learning Representations}.

\bibitem[{Zhang et~al.(2024)Zhang, Tan, Shen, Salem, Backes, Zannettou, and Zhang}]{zhang2024breaking}
Boyang Zhang, Yicong Tan, Yun Shen, Ahmed Salem, Michael Backes, Savvas Zannettou, and Yang Zhang. 2024.
\newblock Breaking agents: Compromising autonomous llm agents through malfunction amplification.
\newblock \emph{arXiv preprint arXiv:2407.20859}.

\bibitem[{Zhou et~al.(2024)Zhou, Liu, Jin, Nie, and Dou}]{zhou2024metacognitive}
Yujia Zhou, Zheng Liu, Jiajie Jin, Jian-Yun Nie, and Zhicheng Dou. 2024.
\newblock Metacognitive retrieval-augmented large language models.
\newblock In \emph{Proceedings of the ACM Web Conference 2024}, pages 1453--1463.

\bibitem[{Zhu et~al.(2024)Zhu, Zhang, and Wang}]{zhu2024language}
Wentao Zhu, Zhining Zhang, and Yizhou Wang. 2024.
\newblock Language models represent beliefs of self and others.
\newblock In \emph{Forty-first International Conference on Machine Learning}.

\bibitem[{Zhu et~al.(2025)Zhu, Qiao, Ou, Deng, Lyu, Shen, Liang, Gu, Chen, and Zhang}]{zhu-etal-2025-knowagent}
Yuqi Zhu, Shuofei Qiao, Yixin Ou, Shumin Deng, Shiwei Lyu, Yue Shen, Lei Liang, Jinjie Gu, Huajun Chen, and Ningyu Zhang. 2025.
\newblock {K}now{A}gent: Knowledge-augmented planning for {LLM}-based agents.
\newblock In \emph{Findings of the Association for Computational Linguistics: NAACL 2025}, pages 3709--3732.

\bibitem[{Zhuo et~al.(2025)Zhuo, Chien, Chim, Hu, Yu, Widyasari, Yusuf, Zhan, He, Paul, Brunner, GONG, Hoang, Zebaze, Hong, Li, Kaddour, Xu, Zhang, Yadav, Jain, Gu, Cheng, Liu, Liu, Wang, Lo, Hui, Muennighoff, Fried, Du, de~Vries, and Werra}]{zhuo2025bigcodebench}
Terry~Yue Zhuo, Vu~Minh Chien, Jenny Chim, Han Hu, Wenhao Yu, Ratnadira Widyasari, Imam Nur~Bani Yusuf, Haolan Zhan, Junda He, Indraneil Paul, Simon Brunner, Chen GONG, James Hoang, Armel~Randy Zebaze, Xiaoheng Hong, Wen-Ding Li, Jean Kaddour, Ming Xu, Zhihan Zhang, and 14 others. 2025.
\newblock Bigcodebench: Benchmarking code generation with diverse function calls and complex instructions.
\newblock In \emph{The Thirteenth International Conference on Learning Representations}.

\bibitem[{Zou et~al.(2023)Zou, Phan, Chen, Campbell, Guo, Ren, Pan, Yin, Mazeika, Dombrowski et~al.}]{zou2023representation}
Andy Zou, Long Phan, Sarah Chen, James Campbell, Phillip Guo, Richard Ren, Alexander Pan, Xuwang Yin, Mantas Mazeika, Ann-Kathrin Dombrowski, and 1 others. 2023.
\newblock Representation engineering: A top-down approach to ai transparency.
\newblock \emph{arXiv preprint arXiv:2310.01405}.

\end{thebibliography}
\clearpage
\appendix
\section{Theoretical Landscape}
\label{appendix:foundations}
In this section, we provide a more comprehensive introduction to consciousness theories. Following \citet{block1995confusion}, we classify contemporary theories of consciousness into three categories: \emph{phenomenal consciousness}, \emph{access consciousness}, and \emph{hybrid theories}. \emph{Hybrid theories} integrate both phenomenal and access aspects, arguing that neither is solely sufficient to explain consciousness.\footnote{This classification is not strictly exclusive; theory placement can vary based on interpretation.}

\subsection{Phenomenal Consciousness}

\textbf{Recurrent processing theory} (RPT) posits that recurrent (or feedback) processing within neural circuits is both necessary and sufficient for consciousness \citep{lamme2000distinct,lamme2010how}. RPT attributes conscious perception to the interaction of higher- and lower-level cortical areas, which results in sustained recurrent processing.
\textbf{Integrated information theory} (IIT) proposes that the degree of conscious experience corresponds to the extent of integrated information $\Phi$ within a system \citep{tononi2004information,tononi2015integrated}.
\textbf{Embodiment theory} (ET) challenges mind-brain dualism \citep{descartes1985meditations}, arguing instead that consciousness is fundamentally linked to the organism's body and environmental \citep{gallagher2005how,gallagher2021phenomenological}. Proponents suggest that embodiment can provide crucial constraints and integrate informational processing, thereby giving rise to genuinely conscious experience.


\subsection{Access Consciousness}
\textbf{Global workspace theory} (GWT) likens consciousness to a central ``stage'' where selective information is shared across multiple specialized processors responsible for perception, memory, emotion, and related functions \citep{baars1988cognitive,dehaene1998neuronal,dehaene2001towards,dehaene2014consciousness}. GWT relies heavily on contrastive analysis, which compares neural activity during conscious versus unconscious processing \citep{dehaene2014consciousness, dehaene2017consciousness,mashour2020conscious}.
\textbf{C0-C1-C2 framework} distinguishes consciousness into three levels: unconscious computations (C0), global information accessibility for report and decision-making (C1), and metacognitive self-monitoring (C2), offering a taxonomy to disentangle often-conflated processes \citep{dehaene2017what}. The framework bypasses the issue of qualia, offering a pragmatic structure for empirical study \citep{birch2022search,chen2024imitation}. 
\textbf{Attention schema theory} (AST) proposes that consciousness arises from the brain’s schematic model of its attentional processes. In this view, consciousness evolves as a simplified internal model of attention, which enhances both the endogenous control of attention and social cognition by attributing attentional states to others \citep{graziano2015attention,graziano2020toward,graziano2020rethinking}.


\subsection{Hybrid Theories}
\textbf{Higher-order theory} (HOT) posits that a mental state becomes conscious only when represented by a distinct, higher-order mental state \citep{rosenthal2005consciousness}. In this view, first-order states represent the external world, while higher-order states are meta-level reflections on those first-order states.
\textbf{Predictive processing} (PP) maintains that the brain operates as a hierarchical prediction machine. It continuously generates top-down predictions about sensory input and updates these predictions based on bottom-up prediction errors \citep{friston2010free, clark2013whatever,seth2022theories}.

\end{document}